\documentclass[10pt,twocolumn,letterpaper]{article}

\usepackage{cvpr}              

\usepackage{xcolor}

\definecolor{myred}{RGB}{255 0 0} 
\definecolor{mygreen2}{RGB}{0 176 80} 
\definecolor{mygreen3}{RGB}{80,99,43} 
\definecolor{mygray}{gray}{.9}

\makeatletter
\newcommand{\thickhline}{%
    \noalign {\ifnum 0=`}\fi \hrule height 1pt
    \futurelet \reserved@a \@xhline
}

\usepackage{adjustbox}
\usepackage{pifont}

\usepackage{multirow}
\usepackage{xcolor}
\usepackage{colortbl}
\usepackage{booktabs}

\usepackage{algorithm}
\usepackage{algorithmic}
\usepackage{newfloat}
\usepackage{listings}
\usepackage{wrapfig}
\usepackage{adjustbox}

\definecolor{cvprblue}{rgb}{0.21,0.49,0.74}
\usepackage[pagebackref,breaklinks,colorlinks,allcolors=cvprblue]{hyperref}

\def\paperID{***} 
\def\confName{CVPR}
\def\confYear{2026}

\title{A-SelecT: Automatic Timestep Selection for \\ Diffusion Transformer Representation Learning}

\author{
\textbf{Changyu Liu\textsuperscript{1}}, 
 \textbf{James Chenhao Liang\textsuperscript{2}},
 \textbf{Wenhao Yang\textsuperscript{3}},
 \textbf{Yiming Cui\textsuperscript{4}},
 \textbf{Jinghao Yang\textsuperscript{5}},\\
 \textbf{Tianyang Wang\textsuperscript{6}},
 \textbf{Qifan Wang\textsuperscript{7}},
 \textbf{Dongfang Liu\textsuperscript{8}},
 \textbf{Cheng Han\textsuperscript{1\thanks{Corresponding author}}}
\\
 \textsuperscript{1}University of Missouri--Kansas City,
 \textsuperscript{2}U. S. Naval Research Laboratory,
 \textsuperscript{3}Lamar University,\\
 \textsuperscript{4}	University of Florida,
 \textsuperscript{5}University of Texas Rio Grande Valley,
  \textsuperscript{6}University of Alabama at Birmingham,\\
 \textsuperscript{7}Meta AI,
 \textsuperscript{8}Rochester Institute of Technology\\
 {{\tt\small cldb5@umkc.edu, james.c.liang.civ@us.navy.mil, wyang2@lamar.edu, }} \\
{{\tt\small cuiyiminghit@gmail.com, jinghao.yang@utrgv.edu, tw2@uab.edu, wqfcr@meta.com,}}\\
{{\tt\small  dongfang.liu@rit.edu, chk9k@umsystem.edu}}
}

\begin{document}
\maketitle
\begin{abstract}
Diffusion models have significantly reshaped the field of generative artificial intelligence and are now increasingly explored for their capacity in discriminative representation learning.
Diffusion Transformer (DiT) has recently gained attention as a promising alternative to conventional U-Net-based diffusion models, demonstrating a promising avenue for downstream discriminative tasks via generative pre-training.
However, its current training efficiency and representational capacity remain largely constrained due to the inadequate timestep searching and insufficient exploitation of DiT-specific feature representations.
In light of this view, we introduce \underline{A}utomatically \underline{Selec}ted \underline{T}imestep (A-SelecT) that dynamically pinpoints DiT's most information-rich timestep from the selected transformer feature in a single run, eliminating the need for both computationally intensive exhaustive timestep searching and suboptimal discriminative feature selection. Extensive experiments on classification and segmentation benchmarks demonstrate that DiT, empowered by A-SelecT, surpasses all prior diffusion-based attempts efficiently and effectively.
\end{abstract}    
\section{Introduction}
\label{sec:intro}
\begin{figure}[t!]
    \centering
    \includegraphics[width=0.49\textwidth]{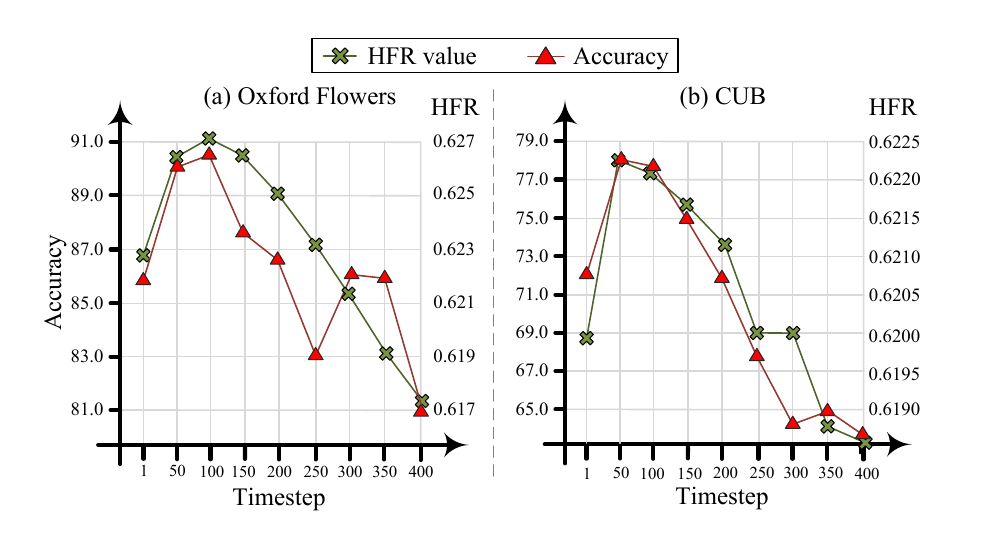}
    \vspace{-8mm}
    \caption{\textbf{A Preliminary Study} on the impact of the High-Frequency Ratio (HFR) with classification performance on Oxford Flowers (a) and CUB (b). The green curve represents the HFR values, and the red curve are classification accuracies. We have two key observations: \textbf{I.} HFR values exhibit a positive correlation with classification accuracies. \textbf{II.} 
    The highest classification accuracy is achieved when the HFR value reaches its maximum. More results in Appendix \S S2 
    }
    \vspace{-5mm}
    \label{fig:HFR}
\end{figure}
In computer vision, representation learning is critical for extracting robust and discriminative features from raw visual data~\cite{zhang2018network}. For decades, Convolutional Neural Networks (CNNs)~\cite{lecun2002gradient, he2016deep, krizhevsky2012imagenet, simonyan2014very} and Vision Transformers (ViTs)~\cite{dosovitskiy2020image, liu2022swin, bao2021beit, caron2021emerging} have served as the foundational architectures for tasks like image classification and semantic segmentation. Diffusion models have recently emerged as a potent alternative for representation learning through generative pre-training. Among these, Diffusion Transformer (DiT)~\cite{peebles2023scalable} has demonstrated remarkable scalability and superior performance in image generation. This success positions DiT as a highly promising candidate for extracting discriminative features via generative pre-training, directly challenging the long-standing dominance of traditional discriminative models in feature extraction tasks.

Despite DiT's promising avenue for discriminative representation learning, two critical challenges notably impede its effectiveness as a feature extractor: \textit{\textbf{\ding{182} Inadequate Timestep Searching.}} Identifying the optimal denoising timestep for extracting the most informative features across potentially hundreds of steps remains a non-trivial and often computationally intensive task; and \textit{\textbf{\ding{183} Insufficient Representation Selection.}}
 The representational quality exhibits variation across transformer blocks, and identifying which specific components within the target block yield the most discriminative features remains a DiT-specific challenge that is yet to be comprehensively investigated.

To address these fundamental challenges, we propose 
a novel framework, \underline{A}utomatically \underline{Selec}ted \underline{T}imestep (A-SelecT), designed to enable DiT as an efficient and effective representation feature extractor.
Specifically, to solve challenge \textbf{\ding{182}}, our approach first introduces the High-Frequency Ratio (HFR),
a principled method designed to dynamically identify the most informative timestep in a single pass. 
Our designed HFR, based on extensive observations and experiments, is always positively correlated to stronger DiT discriminative behavior (see Fig.~\ref{fig:HFR}). To solve challenge \textbf{\ding{183}}, we perform an in-depth analysis of DiT transformer block's components, examining their representational quality. 

Our proposed method offers several key advantages that significantly advance the state of the art of diffusion attempts. First, A-SelecT dramatically reduces computational overhead (\ie, ${\sim}21\times$) by reducing the reliance on expensive traversal search or subjective manual selections. For the first time, A-SelecT can automatically select the optimal timestep for feature extraction in a single trial via HFR (see \S\ref{sec:HFR}). 
Second, our comprehensive analysis of the inner transformer design of DiT (see \S\ref{sec:diag_experiments})
ensures that the selected features are empirically optimal for diverse representation learning downstream tasks (see \S\ref{sec:main_result}) --- achieving 82.5\% on FGVC and 45.0\% on ADE20K. 
Altogether, 
our proposed A-SelecT firmly establishes DiT as a 
strong alternative to traditional CNN and ViT feature extractors.
\begin{figure*}[t!]
    \centering
    \includegraphics[width=1.0\textwidth]{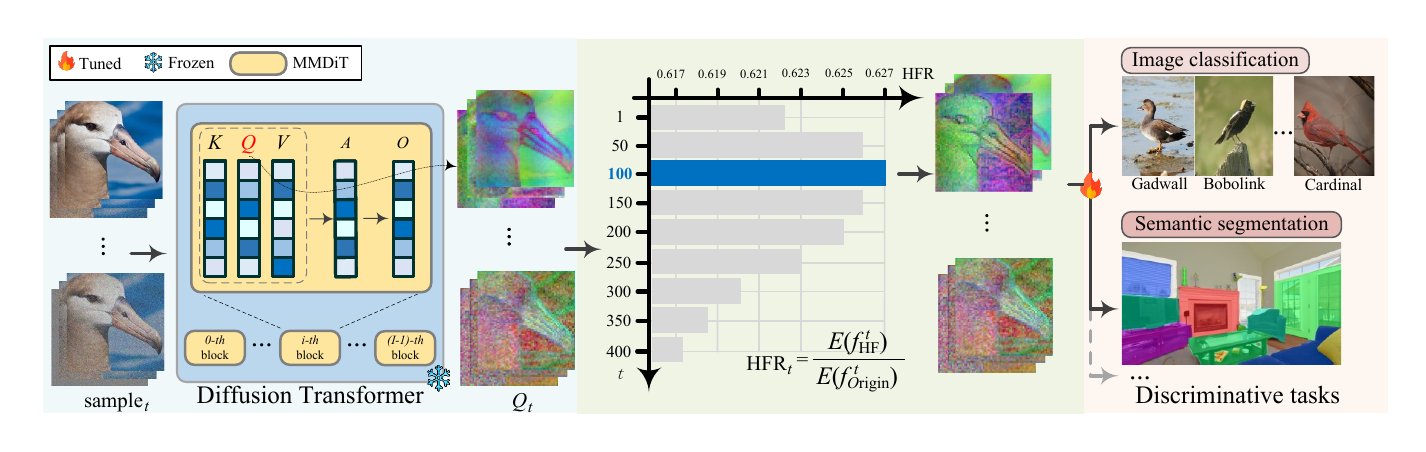}
    \vspace{-6.8mm}
    \caption{\textbf{Overview of \underline{A}utomatically \underline{Selec}ted \underline{T}imestep (A-SelecT).} A-SelecT begins by simulating \( \text{sample}_{t} \) at timestep \(t\) (see Eq.~\ref{eq:forward}). This \( \text{sample}_{t} \) is then processed through the diffusion backbone to extract the query feature \(Q_t\) at each timestep \(t\). Upon obtaining all \(Q_t\), their HFR are calculated. The timestep exhibiting the highest average HFR is subsequently selected for feature extraction. The query feature extracted at this optimal timestep $\hat{t}$ is then fed into targeted discriminative tasks.
    }
    \vspace{-3mm}
    \label{fig:overview}
\end{figure*}
\section{Related Work}
\label{sec:rl}
\subsection{Fast Fourier Transform in Vision}
Fast Fourier Transform (FFT)~\cite{nussbaumer1981fast,almeida2002fractional} is an effective algorithm for computing the Discrete Fourier Transform of a sequence, enabling the transformation of signals from the temporal or spatial domain into the frequency domain. FFT is widely employed in applications such as audio signal analysis~\cite{megias2010efficient, cancela2009efficient, wu2021environmental, hershey2017cnn}, radar signal analysis~\cite{brandwood2012fourier, sifuzzaman2009application, xu2011radon, heo2021fpga}, and image processing~\cite{uzun2005fpga, ell2006hypercomplex,hinman1984short, beaudoin2002accurate}. In computer vision, by revealing the underlying frequency components, which are often more informative and interpretable than raw signals, FFT plays a foundational role in numerous tasks, including deblurring~\cite{kong2023efficient, mao2023intriguing, jiang2024fast, zhou2024seeing}, super-resolution~\cite{fuoli2021fourier, cheng2019fft,li2018frequency}, and texture analysis~\cite{mardani2020neural, xiong2006texture, zeng2024visual}. 
Some recent works~\cite{wang2022antioversmooth, kim2025autoregressive, anagnostidis2025flexidit, patro2025spectformer, dai2024freqformer} explore the use of FFT in analyzing the behavior of vision transformers, revealing that they often exhibit low-pass filtering behavior, thereby resulting in struggling to retain 
high-frequency information. 
High-frequency information, however, has been demonstrated to be crucial for capturing fine-grained details in representation learning~\cite{fuoli2021fourier, wang2022antioversmooth, zeng2024visual, kong2023efficient}. 
Regarding DiT, the denoising process introduces timestep-dependent noise levels, which can directly influence the amount of high-frequency information preserved in features.
Recognizing the critical role of high-frequency information in representation learning, we tailor HFR, a dedicated approach that identifies and selects high-frequency-rich features from DiT, thereby improving its effectiveness as a feature extractor for 
discriminative tasks.

\subsection{Diffusion Models for Feature Extraction}\label{sec:rl_discrimination}
Diffusion models have recently demonstrated their capabilities in representation learning. Their generative pre-trained models are generally utilized in three primary ways: \textit{Training-free}~\cite{li2023your, chen2024your} applies Bayes’ theorem without requiring additional training. However, they suffer from impractically slow inference. \textit{Fine-tuning}~\cite{garcia2024fine, ke2024repurposing, chen2024deconstructing, he2022masked, han2022card,stracke2025cleandift, amit2021segdiff, ji2023ddp} involves fully updating the backbone of generative pre-trained diffusion models, which is naturally computationally expensive and extremely time-consuming. \textit{Feature extraction}~\cite{mukhopadhyay2024text, meng2025not, zavadski2024primedepth, luo2023diffusion, tang2023emergent,garcia2025fine,baranchuk2021label}, on the other hand, employs features extracted from pre-trained diffusion models to train a lightweight downstream network, providing a way more effective alternative on discriminative performance. 
Though promising, this way suffers from extended training overhead and less effective representation learning.

For training overhead, identifying the most informative timestep for feature extraction from diffusion models remains challenging, as the denoising process involves numerous steps. Straightforward approaches include \textit{traversal search}~\cite{zhang2023tale, li2024sd4match, zhang2024telling} and \textit{fixed search}~\cite{zavadski2024primedepth, ji2023ddp}, which brute-force trains a downstream network for each individual timestep, and extracts features solely from the final timestep (\ie, fixed), respectively. While \textit{traversal search} is computationally impractical, studies~\cite{mukhopadhyay2024text, mukhopadhyay2023diffusion} show that \textit{fixed search} also leads to suboptimal performance. 
Less effective representation learning is another critical challenge.  
Attempts such as \textit{feature visualization}~\cite{meng2025not} rely only on manual inspection of feature maps, which is subjective and impractical. Shown in \S\ref{subsec:dit_with_feature_vis}, we confirm that human judgments in DiT are inconsistent,
leading to unsatisfying performance. 
Other attempts (\eg, Denoising Diffusion Autoencoders~\cite{ddae2023}, Latent Denoising Autoencoder~\cite{chen2024deconstructing}, REPresentation Alignment~\cite{yu2024representation}) directly extract features from
the DiT layer-wise outputs, without considering the detailed inner design of DiT-specific transformer architecture, ultimately resulting in less effective representation learning.

Acknowledging the major insufficiency on the two directions, we utilize HFR as a reliable high-frequency indicator for fast and accurate automatic timestep selection, and
offer an in-depth examination of the representational dynamics within DiT transformer blocks, altogether establishing DiT as an efficient and effective feature extractor.
\section{Method}
\label{sec:method}
We begin by introducing diffusion model fundamentals in \S\ref{sec:prelim}.  Subsequently, in \S\ref{sec:problem-form}, we formally define the timestep selection problem for DiT representation learning. To address this problem, \S\ref{sec:HFR} introduces High-Frequency Ratio (HFR). As detailed in \S\ref{sec:opt_sel}, we automatically assess the optimality of the timestep selection for the discriminative feature in only a single trial via HFR. Hence, we term
our method as \underline{A}utomatically \underline{Selec}ted \underline{T}imestep (A-SelecT). Finally \S\ref{sec:whyhfr} provides theoretical insights into why HFR functions as a reliable and principled indicator.

\subsection{Preliminaries}\label{sec:prelim}
Diffusion models, such as Stable Diffusion 3.5~\cite{esser2024scaling} and  EDM~\cite{Karras2022edm}, are fundamentally conceptualized through the framework of ordinary differential equations~\cite{chen2018neural, song2020score}, encapsulating the forward diffusion process and the backward denoising process. Specifically, in the forward process, the model constructs a noised representation \( z_t \) by interpolating between the initial data representation and a stochastic noise component, expressed as:
\begin{equation}
z_t = \alpha_t \cdot \epsilon + (1 - \alpha_t) \cdot z_0,
    \label{eq:forward}
\end{equation}
where \( z_0 \) is the original representation, \( \epsilon \sim \mathcal{N}(0, I) \) is the Gaussian noise introduced into \( z_0 \), and \( \alpha_t \in [0, 1] \) is a time-dependent scalar that controls the temperature of the noise.

The backward process, also known as the sampling or denoising phase, progressively removes the noise and reconstructs the original data representation iteratively as:
\begin{equation}
z_{t-1} = z_t + \Delta \alpha \cdot v_\theta(z_t, \alpha_t),
    \label{eq:backward}
\end{equation}
where \( \Delta \alpha  \) is calculated by $\alpha_{t} - \alpha_{t-1}$ and denotes the step size. \( v_\theta(z_t, \alpha_t) \)
is a velocity vector that indicates the denoising direction. It progressively guides \( z_t \) towards the target data distribution. By integrating the velocity field over time, the model reverses the forward diffusion process, transforming the noise into a reconstructed representation.

\subsection{Problem Formulation}\label{sec:problem-form}
This study investigates the structural framework of Stable Diffusion (SD) 3.5~\cite{esser2024scaling}, with a focus on identifying optimal selection of DiT features for learning discriminative representation. Specifically, our objective is to determine the timestep \( \hat{t} \in [1, T]\)
at which the selected feature from DiT exhibits optimal representation learning properties.

However, accurately identifying \( \hat{t} \) poses several key challenges. \textit{First}, the number of timesteps is large (\ie, $T$ steps) and the optimal timestep is pretty flexible. For different discriminative datasets, their optimal timesteps can exhibit significant variation (see Fig.~\ref{fig:HFR}). \textit{Second}, the overall computational cost of finding the maximum performance at the optimal timestep becomes noticeably high when conducting a brute-force search (see \S \ref{subsec:dit_with_traversal_search}). \textit{Third}, the alternative approach to visualizing features at each step to assess their discriminative quality manually is highly subjective and heavily relies on human judgments (see \S \ref{subsec:dit_with_feature_vis}).

In practice, the diffusion models comprise $I$ blocks, each configured as a multimodal DiT (MMDiT) block. Unlike the U-Net ~\cite{ronneberger2015u} architecture, where the input and output dimensions vary~\cite{sun2024unveiling, wang2025seedvr}, each block in the MMDiT maintains consistent dimensions throughout, ensuring uniformity across all processing stages. 
This distinctive design distinguishes DiT from conventional U-Net-based diffusion models in structural composition, thereby necessitating distinct approaches to representation learning.

Within the $i$-th MMDiT block ($i \in  [0, I-1]$), the attention layer is a critical component, processing Query \( Q^{i} \), Key \( K^{i} \), and Value \( V^{i} \) as inputs. Our analysis, detailed in \S\ref{sec:diag_experiments}, identifies the optimality, where both the output from attention layers (\ie, \( A^{i} \)) and the representations derived from the MMDiT block (\ie,  \( O^{i} \)) could be considered as potential candidates for discriminative feature extraction. Without loss of generality, we comprehensively study the optimal candidates in \S\ref{sec:opt_sel}.
We aim to analyze \( Q^{i} \) extracted from a total of $T$ timesteps in the backward process. 
For clarity, we denote  \( Q^{i} \) as \( Q\) for the remainder of this study.

\subsection{High-Frequency Ratio}
\label{sec:HFR}

To tackle these challenges, we observe and introduce the High-Frequency Ratio (HFR), a novel quantitative metric designed to identify the level of informative feature at timestep $t \in [1, T]$.
HFR quantifies the extent to which the high-frequency information from the $t$-th step feature contributes to the overall feature representation. A higher HFR signifies a more favorable capability of the feature to capture the high-frequency information, and vice versa. 

\begin{figure}[t]
    \centering
    \includegraphics[width=0.4\textwidth]{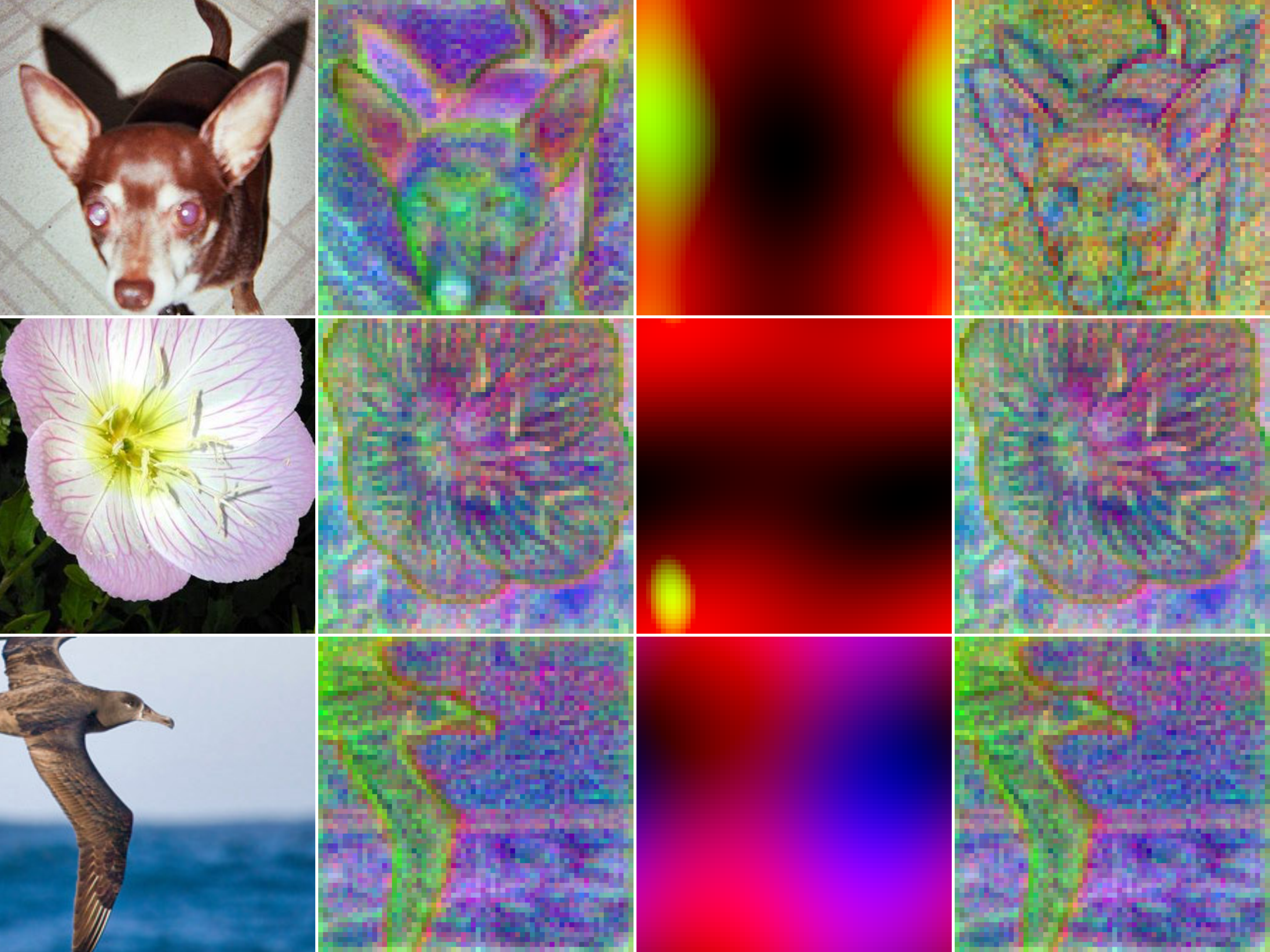}
    \put(-189,-9){\small Image}
    \put(-144,-9) { \small Original}
    \put(-92,-9){\small Low-freq}
    \put(-42,-9){\small High-freq}
    \vspace{-1.5mm}
    \caption{\textbf{Visualizations of High-frequency $vs.$ Low-frequency Information.} We present a decomposition of the original features extracted from SD 3.5 into components that exclusively contain high-frequency and low-frequency information. The second column is the original features extracted from the model. As seen, the high-frequency features are shown to contain more discriminative information (\ie, edge, texture, corner information from the black footed albatross can be clearly preserved) than their low-frequency counterparts. Inspired by this, we design HFR to assess the significance of high-frequency information (see \S\ref{sec:HFR}).}
    \vspace{-4mm}
    \label{fig:high_low_vis}
\end{figure}
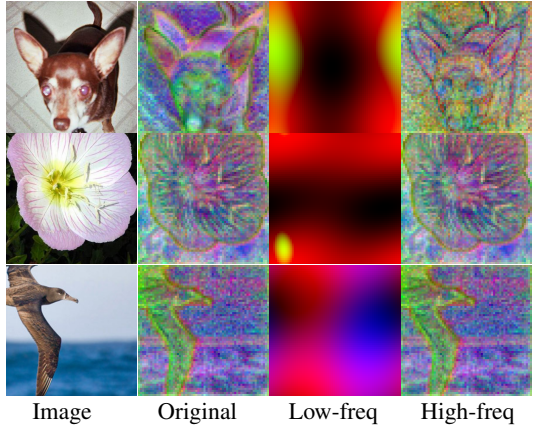

\noindent\textbf{\textit{Observation of High-Frequency Components.}} HFR is inspired by our observation that high-frequency information, which includes fine image details such as edges, textures, and corners, faithfully possesses more discriminative power. We thus employ a Gaussian high-pass filter to separate the original diffusion features into components containing solely high-frequency information and those with solely low-frequency information (see Fig.~\ref{fig:high_low_vis}). This separation allows us to visually compare the two types of features, clearly demonstrating that high-frequency features contain more discriminative details for downstream representation learning than their low-frequency counterparts. The preliminary results strongly support our observation (see Fig.~\ref{fig:HFR}), indicating that timesteps characterized by features with higher HFR values (\ie, greater capacity for high-frequency information) 
is positively related to superior discriminative performance across datasets.

\noindent\textbf{\textit{Definition of the High-Frequency Ratio.}} Based on this insight, we define the High-Frequency Ratio at timestep $t$ as:
\begin{equation}
\text{HFR}_{t} = \frac{E(f_{\text{HF}}^{t})}{E(f_{\text{Origin}}^{t})}.
    \label{eq:HFR}
\end{equation}
Here \( E(\cdot) \) is the summation of the squared magnitudes of its constituent values and represents the energy of the feature representation. At timestep $t$,
\( f_{\text{Origin}}^{t} \) is the original feature extracted from the diffusion model. \( f_{\text{HF}}^{t} \) is the high-frequency component extracted from \( f_{\text{Origin}}^{t} \) as:
\begin{equation}
f_{\text{HF}}^{t} = F^{-1}(G \odot F(f_{\text{Origin}}^{t})),
    \label{eq:HFfeature}
\end{equation}
where \( F(\cdot) \) and \( F^{-1}(\cdot) \) denote the Fast Fourier Transformation (FFT) and its inverse, respectively. \( \odot \) is the Hadamard product, and \( G(\cdot) \) is a Gaussian high-pass filter. \\
\noindent\textbf{\textit{Preliminary Results.}} To assess the efficacy of the HFR, we conducted preliminary experiments on image classification, specifically on CUB~\cite{wah2011caltech} and Oxford Flowers~\cite{nilsback2008automated} datasets over $T$ timesteps (\ie, $T=1000$). For each timestep $t$, we train a separate downstream classifier, report its performance, and compute its HFR. 
The results, as shown in Fig.~\ref{fig:HFR}, clearly demonstrate a positive correlation between HFR values and the classification performance: the highest accuracy is achieved at the timestep where the HFR is maximized. We report consistent results on other datasets and tasks (see Appendix \S S2), 
clearly showing the efficacy of HFR as a robust indicator of discriminative feature quality. 
To demonstrate HFR's generalization, we further extend it on different DiT models (see Appendix \S S14), where the results align with our current observations.

\subsection{Automatically Selected Timestep (A-SelecT)}\label{sec:opt_sel}

Leveraging HFR as the frequency-aware criterion, we can automatically select the optimal timestep for discrimination.
Hence, our objective thus turns into extracting the most informative feature for downstream representation learning at timestep t, which aims to identify the timestep that yields the highest HFR value among all potential candidate features.
We name the automatic searching pipeline as Automatically Selected Timestep (A-SelecT) (see Fig.~\ref{fig:overview}). 
As experimentally shown in \S\ref{sec:diag_experiments}, both \(Q\) and \(V\) yield strong results on downstream representation learning; however, \(Q\) averagely achieves superior discriminative results compared to \( K \) and \( V \). 
We thus apply HFR on query feature \(Q\) at timestep \(t\), denoted as \(Q_t\) for consistency in our study.

To compute the $\text{HFR}_{t}$ for the query feature \(Q_t\), the conventional method typically involves progressing through the backward diffusion process to obtain \( \text{sample}_{t-1} \) from a noise, and subsequently feeding this \( \text{sample}_{t-1} \) into the diffusion backbone to extract \(Q_t\). However, this is notably time-consuming due to the extensive sampling. Following~\cite{mukhopadhyay2024text}, we instead employ the forward process (\ie, Eq.~\ref{eq:forward}) to simulate \( \text{sample}_{t-1} \) by combining a single input image from training data with a noise sampled from a standard Gaussian distribution \(N(0, I)\). In this manner, the computational overhead is significantly reduced by bypassing the time-intensive backward process (\ie, ${\sim}100\times$ faster).
Once \(\text{sample}_{t-1} \) is simulated, we can directly extract \(Q_t\) from diffusion backbone. Next, this feature from a single input is utilized to calculate its $\text{HFR}_{t}$ following Eq.~\ref{eq:HFR}. To perform an evaluation across the dataset, we compute the HFR for each image and subsequently get the average HFR as: $\tilde{\text{HFR}}_t=\frac{1}{N}\sum_{i=0}^{N}\text{HFR}_{t}$, where $N$ is the dataset length. 
Lastly, we are able to automatically choose the highest $\tilde{\text{HFR}}_t$ value by spanning across all $T$ as:
\begin{equation}
t' = \arg\max_{t \in [1, T]} \tilde{\text{HFR}}_t.
\label{eq:A-SelecT}
\end{equation}

Here, $t'$ is the timestep found by A-SelecT for extracting discriminative features from the diffusion model. We experimentally show that $t'$ is always equal to $\hat{t}$ under different downstream representation learning tasks, indicating the effectiveness of A-SelecT (see \S\ref{sec:main_result} and \S\ref{sec:diag_experiments}).

\subsection{Understanding HFR with Fisher Score}\label{sec:whyhfr}

To further explore why HFR acts as a robust indicator for timestep selection, 
we analyze its relationship with the \emph{Fisher Score}: 
a classical and widely used criterion for evaluating the discriminative power of features in statistical learning~\cite{he2005laplacian, de2021scores, aksu2018intrusion,gu2012generalized}. It measures how effectively features distinguish between classes by comparing the variability across different classes to the variability within each class, where higher Fisher Scores indicate better class separability and stronger discriminative capability. Formally, given a dataset of feature embeddings $\{x_i\}_{i=1}^{N}$ extracted from a trained model and their corresponding class labels $y_i \in \{1, \dots, C\}$. 
The overall Fisher Score is defined as:
\begin{equation}
J = \frac{\operatorname{tr}(S_b)}{\operatorname{tr}(S_w)},
\end{equation}
where $\operatorname{tr}(\cdot)$ denotes the matrix trace operator. $S_b$ and $S_w$ represent the between-class and within-class variations, respectively. They are defined as:
\begin{equation}
S_w = \sum_{k=1}^{C} \sum_{x_i \in \mathcal{D}_k} (x_i - \mu_k)(x_i - \mu_k)^\top, 
\end{equation}
\begin{equation}
S_b = \sum_{k=1}^{C} n_k (\mu_k - \mu)(\mu_k - \mu)^\top,
\end{equation}
where $\mu_k \in \mathbb{R}^d$ represents the mean embedding of class $k$, $\mu$ the global mean of all samples, $n_k$ the number of samples in class $k$, and $\mathcal{D}_k$ the set of samples belonging to class $k$. The ratio $J$
quantifies the balance between inter-class separation and intra-class compactness, offering a principled and reliable measure of feature discriminability.

\begin{figure}[t!]
    \centering
    \vspace{1mm}
    \includegraphics[width=0.5\textwidth]
    {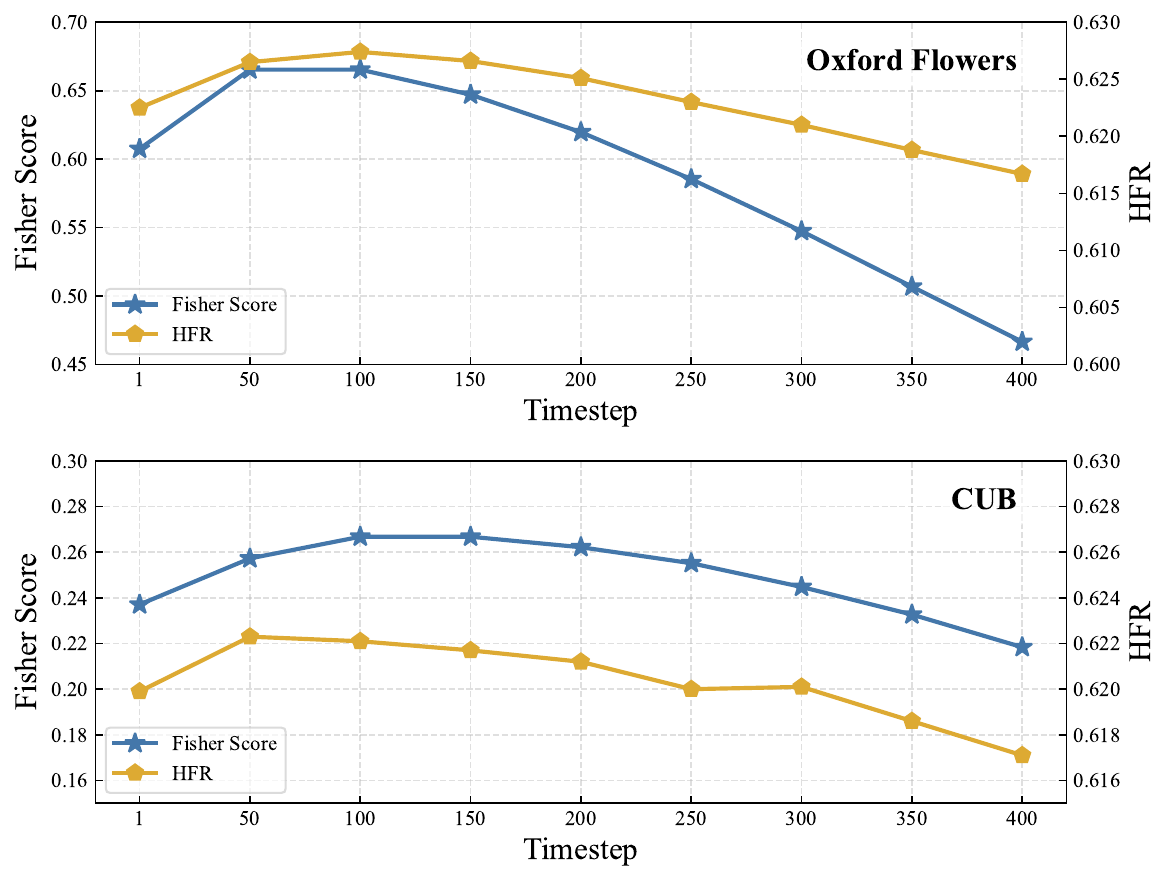}
    \caption{\textbf{Comparison of HFR and Fisher Score} across time- steps on Oxford Flowers (top) and CUB (bottom). They show strong alignment, indicating that HFR captures discriminative characteristics consistent with the Fisher Score and serves as a reliable label-free indicator of feature separability.}
    \vspace{-3mm}
    \label{fig:HFRvsFS}
\end{figure}

We then compute the Fisher Scores of features extracted at different timesteps and compare them with our proposed 
HFR (see Fig.~\ref{fig:HFRvsFS}).
They exhibit highly aligned trends, indicating that HFR quantifies feature discriminability in a manner consistent with the established statistical principles and providing theoretical justification for its effectiveness. 
The reason we do not directly adopt Fisher Score as the criterion is that it requires ground truth label information, which is unavailable during testing. As it is infeasible to compute, HFR stands as the only effective alternative.
To sum up, we prove that HFR is not merely an empirical indicator but a theoretically grounded criterion for identifying the most discriminative timestep in A-SelecT. Additional details on Fisher Score are provided in Appendix \S S13. 

\section{Experiment}
\label{sec:exp}
We present a comprehensive analysis of A-SelecT through a series of traditional representation learning tasks, including image classification and semantic segmentation. We detail the datasets utilized, outline the implementation specifics, and compare our approach against state-of-the-art baselines to substantiate its efficacy in this section as well. More experiments are provided in Appendix \S S1-\S S14.

\begin{table*}[ht]
\vspace{-0.5cm}

\begin{adjustbox}{width=0.915\width,center}
\begin{tabular}{l||cccccc|c} 
\hline \thickhline
\rowcolor{mygray}
Method &  Aircraft&  Stanford Cars & CUB & Stanford Dogs &  Oxford Flowers &  NABirds & Mean \\
\hline \hline

$\text{ResNet-50}^{\star}$ ~\cite{he2016deep}   &63.8\%  &68.8\% &64.0\% &82.3\% &82.5\% &54.2\%  & 69.3\%\\
SimCLR ~\cite{chen2020simple} &46.4\%  &40.9\%  & 43.9\% &64.2\%  & 86.2\% &35.9\% & 52.9\%\\
SwAV ~\cite{caron2020unsupervised} &58.2\%  & 57.6\% & 63.8\% & 74.9\% & \textbf{91.8\%} & 54.1\%& 66.7\%\\
MAGE ~\cite{li2023mage} &67.6\%  &73.4\%  &\underline{77.7}\%  &\textbf{88.2\%}  & 86.8\% &\underline{76.7}\%  &\underline{78.4\%} \\
GD ~\cite{mukhopadhyay2023diffusion}  &57.2\%  &24.5\%  &32.8\%  &58.3\%  &78.4\%  &20.1\%  &45.2\% \\
DifFeed ~\cite{mukhopadhyay2024text} &73.2\%  &82.7\%  & 71.0\% & 81.3\%&  90.1\%& 70.7\% &78.2\%\\
SDXL ~\cite{meng2025not}    &\underline{73.7}\% &\underline{83.2}\% &71.9\% &82.1\% &87.5\% &71.6\% &78.3\% \\
\hline 
\rowcolor{mygray}
Ours & \textbf{77.5\%} & \textbf{86.1\%} & \textbf{78.6\%}  & \underline{83.5\%} & \underline{90.6\%} & \textbf{78.4\%} & \textbf{82.5\%}\\
\hline
\end{tabular}
\end{adjustbox}

\caption{\textbf{Image Classification Results on FGVC Benchmark.} The best results are highlighted in \textbf{bold}, and the second best are shown in \underline{underline}. Same for Table \ref{table:table_gfvc}-\ref{table:table_ade20k}. We report top-1 accuracy from 6 baselines and our A-SelecT on FGVC. Our approach achieves the best performance in 4 out of 6 datasets and ranks second in the remaining datasets. A-SelecT consistently outperforms U-Net-based diffusion models (\ie, DifFeed, GD, and SDXL) across all datasets. $^{\star}$ means that ResNet-50 backbone is frozen during representation learning, and applied solely for feature extraction. This setting is consistent with other baselines' design for fairness. See Appendix \S S3 for more details.}
\label{table:table_gfvc}
\end{table*}
\subsection{Experimental Setup}\label{sec:exp setup}
\noindent \textbf{Datasets.} 
Our evaluation comprehensively encompasses both classification and segmentation tasks. For classification, we evaluate on ImageNet dataset~\cite{deng2009imagenet} and Fine-Grained Visual Classification (FGVC) benchmark, including six separate datasets: Caltech-UCSD Birds (CUB)~\cite{wah2011caltech}, Aircraft~\cite{maji2013fine}, Stanford Cars~\cite{krause20133d}, NABirds~\cite{van2015building}, Stanford Dogs~\cite{khosla2011novel}, and Oxford Flowers~\cite{nilsback2008automated}. For the segmentation task, the ADE20K dataset~\cite{zhou2017scene} serves as the benchmark. 

\noindent \textbf{Baselines.} To evaluate the effectiveness of A-SelecT, we compare it with several state-of-the-art approaches related to our study.
For image classification, we include three diffusion-based methods (\ie, DifFeed~\cite{mukhopadhyay2024text}, GD~\cite{mukhopadhyay2023diffusion}, SDXL~\cite{meng2025not}), a GAN-based method (\ie, BigBiGAN~\cite{donahue2019large}), four self-supervised learning methods (\ie, SimCLR~\cite{chen2020simple}, SwAV~\cite{caron2020unsupervised}, MAE~\cite{he2022masked}, MAGE~\cite{li2023mage}) and a supervised baseline (\ie, ResNet-50~\cite{he2016deep}). Note that for fairness, we introduce ResNet-50 as a feature extractor for FGVC, which keeps the model completely frozen. Since ResNet-50 is generally pre-trained on ImageNet, we do not report its result in Table~\ref{table:table_iamgenet} for fair comparison. 
For semantic segmentation, we further include a diffusion-based method (\ie, SDXL-t~\cite{meng2025not}),
and one additional self-supervised learning method (\ie, DreamTeacher~\cite{li2023dreamteacher}).
By benchmarking A-SelecT with these, we assess the effectiveness of our method in extracting high-quality discriminative features and its potential advantages over conventional self-supervised or generative paradigms, more excitingly, surpassing some supervised baselines, revealing a promising future for utilizing generative pre-training models. More results on other DiT models are provided in Appendix \S S14.

\noindent \textbf{Implementation Details.} 
We follow the implementation settings with DifFeed~\cite{mukhopadhyay2024text}, maintaining the pre-trained DiT frozen while only training the downstream discriminative heads (see \S\ref{sec:rl_discrimination}). Architectural level differently, we extract features from Stable Diffusion 3.5 Medium~\cite{esser2024scaling}, which comprises $24$ MMDiT blocks and features a backward process with 1,000 time steps for denoising. Data preprocessing involves normalization using a mean of [0.5, 0.5, 0.5] and a standard deviation of [0.5, 0.5, 0.5], accompanied by random flipping, resizing, and cropping of images, with crop sizes variably selected from the dimensions \{256, 512, 1024\}. 
For optimization, we employ AdamW optimizer~\cite{loshchilov2017decoupled}, regulated by a cosine annealing schedule without warm-up epochs, experimenting with initial learning rates of \{0.001, 0.002, 0.003\}. The batch size is set at $8$. Features are extracted
from layers \{0, 3, 6, 9, 12, 15, 18, 21, 23\}, with the final feature set consistently extracted from the 9-${th}$ block for \textit{all} tasks. Training extends over $28$ epochs for classification tasks and $160K$ iterations for segmentation tasks, following established protocols~\cite{wang2022visual}.

\noindent \textbf{Reproducibility.} A-SelecT is implemented in Pytorch \cite{NEURIPS2019_9015}. Experiments are conducted on NVIDIA A6000 GPUs. Our implementation will be released for reproducibility. We provide the pseudo code 
in Appendix Algorithm 1.

\subsection{Main Results}\label{sec:main_result}

\noindent
\textbf{A-SelecT on FGVC.} 
Table~\ref{table:table_gfvc} reports the top-1 accuracy on FGVC benchmark. 
These results lead to several key observations.
\textit{First}, when compared to SDXL, another diffusion-based discriminative approach, our method consistently delivers superior performance across \textbf{ALL} tasks. For example, our method achieves substantial improvements on NABirds (\ie, \textbf{6.8\%}) and CUB (\ie, \textbf{6.7\%}), respectively. 
\textit{Second}, when compared to other strong baselines, our A-SelecT can get superior performance in \textbf{4 out}
\textbf{of 6} tasks. In the remaining two tasks, we are still able to achieve the second-highest performance (\eg, 90.6\% $vs.$ 91.8\% on Oxford Flowers). 
This clearly demonstrates that DiT can serve as a strong discriminative learner, even when compared to models specifically designed for discrimination. \textit{Third}, leveraging the capabilities of A-SelecT, we could determine the optimal step for selecting the diffusion discriminative feature, ensuring both training efficiency and robust model performance.
More discussions are included in \S\ref{sec:discussion}.

\setlength{\columnsep}{3mm}
\begin{wraptable}{r}{4.4cm}
\setlength{\belowcaptionskip}{-.5cm}

\begin{adjustbox}{width=\linewidth}
\begin{tabular}{l||c} 
\hline \thickhline
\rowcolor{mygray}
Method &    Top-1 Acc\\
\hline \hline

SimCLR ~\cite{chen2020simple}  &69.3\%\\ 
SwAV ~\cite{caron2020unsupervised}   &75.3\%\\ 
MAE ~\cite{he2022masked}   &73.5\%\\ 
MAGE ~\cite{li2023mage}    &\textbf{78.9\%}\\ 
BigBiGAN  ~\cite{donahue2019large}    &60.8\%\\ 
DifFeed  ~\cite{mukhopadhyay2024text}  &77.0\%\\
SDXL ~\cite{meng2025not}    &77.2\%\\
GD  ~\cite{mukhopadhyay2023diffusion}   &71.9\%\\
\hline 
\rowcolor{mygray}
Ours & \underline{78.2\%} \\
\hline
\end{tabular}
\end{adjustbox}
\vspace{-3mm}
\hspace{-5mm}
\caption{\textbf{Image Classification Results on ImageNet~\cite{deng2009imagenet}.}} 
\label{table:table_iamgenet}
\end{wraptable}
\noindent\textbf{A-SelecT on ImageNet.} We further conduct ima- ge classification on ImageNet for completeness. 
As shown in Table~\ref{table:table_iamgenet}, our method achieves \textbf{78.2\%} accuracy, outperforming diffusion-based models DifFeed and SDXL by \textbf{1.2\%} and \textbf{1.0\%}, respecti- vely, while demonstrating 
a compelling improvement over the GAN-based model (\ie, BigBiGAN) by a substantial margin of \textbf{17.4\%}. 
Furthermore, A-SelecT shows superior performance compared to most existing self-supervised learning approaches and yields results comparable to those achieved by MAGE (\ie, 78.2\% $vs.$ 78.9\%), a leading method in self-supervised learning. 
These results are impressive and strengthen 
that, beyond DiT conventional application in image generation, it exhibits great potential in discrimination.

\setlength{\columnsep}{3mm}
\begin{wraptable}[17]{R}{4.35cm}
\setlength{\abovecaptionskip}{2mm}
\vspace{-3mm}
\begin{adjustbox}{width=\linewidth}
\begin{tabular}{l||c} 
\hline \thickhline
\rowcolor{mygray}
Method &    mIoU\\
\hline \hline
ResNet-50 ~\cite{he2016deep}   &40.9\%\\ 
SimCLR ~\cite{chen2020simple}   &39.9\%\\ 
SwAV ~\cite{caron2020unsupervised}  &41.2\%\\ 
MAE (ViT-B) ~\cite{he2022masked}   &40.8\%\\ 
MAE (ViT-L) ~\cite{he2022masked}  &\textbf{45.8\%}\\ 
DreamTeacher ~\cite{li2023dreamteacher}   &42.5\%\\ 
DifFeed ~\cite{mukhopadhyay2024text}  &44.0\%\\
SDXL ~\cite{meng2025not}    &43.5\%\\
$\text{SDXL-t}^{\dag}$ ~\cite{meng2025not}    &\cellcolor{gray!80}{45.7\%}\\
\hline 
\rowcolor{mygray}
Ours   & \underline{45.0\%} \\
\hline
\end{tabular}
\end{adjustbox}
\caption{\textbf{Semantic Segmentation on ADE20K} \cite{zhou2017scene}. $^{\dag}$: SDXL-t uses features from other models (\ie, SD1.5~\cite{rombach2022high} and Playground2~\cite{li2024playground}), which is unfair when compared to other baselines.} 
\vspace{-2mm}
\label{table:table_ade20k}
\end{wraptable}
\noindent \textbf{Results on Semantic Segmentation.}
To further explore the generalizabil-
ity of A-SelecT, we dir- 
ectly apply it to the sema-
ntic segmentation task, ADE20K. 
As shown in Table~\ref{table:table_ade20k}, our method attai-
ins a mean Intersection over Union (mIoU) of 45.0\%, exceeding the DifFeed by \textbf{1.0\%}.
A-SelecT further outperforms the supervised ResNet-50 by
\textbf{4.1\%}, and exceeds the performance of most self-
supervised learning methods. For example, our method achieves \textbf{3.8\%} improvement when compared to SwAV and comparable performance to MAE with ViT-L~\cite{dosovitskiy2020image}, respectively. It is noteworthy that during the training of the segmentation head, we freeze the whole diffusion backbone completely. MAE, on the other hand, necessitates full fine-tuning of the whole backbone with the segmentation head during training. 
In conclusion, A-SelecT in segmentation substantiate the strong potential of DiT in attaining state-of-the-art performance.

\subsection{Diagnostic Experiments}\label{sec:diag_experiments}
\begin{figure}[t!]
    \centering
    \vspace{-10mm}
    \includegraphics[width=0.50\textwidth, trim=110pt 20pt 40pt 40pt, clip ]
    {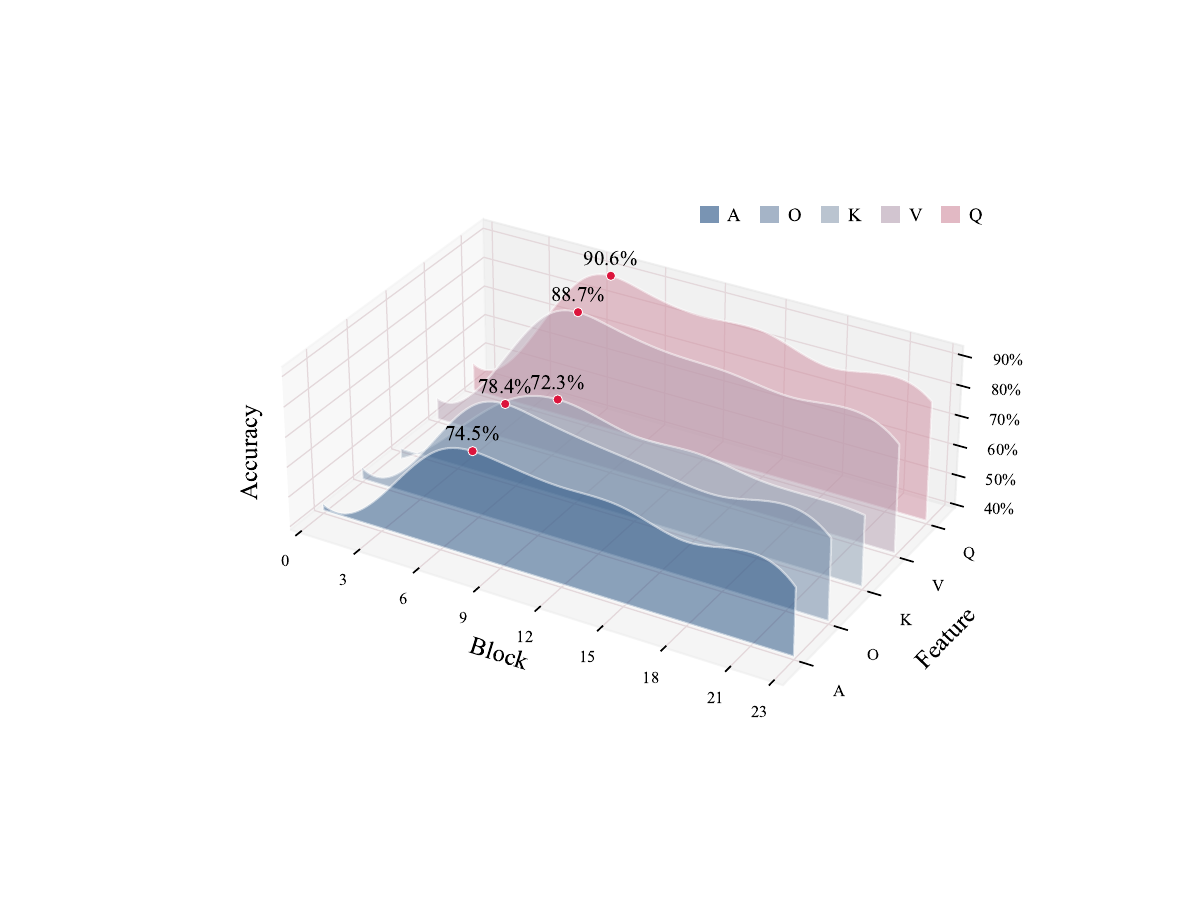}
    \vspace{-17mm}
    \caption{\textbf{Impact of Feature and Block Selection.} We present accuracy across the features $Q$, $K$, $V$, $A$, and $O$ extracted from different transformer blocks on Oxford Flowers. $Q$ and $V$ achieve the highest accuracies (90.6\% and 88.7\%, respectively), while $A$, $O$, and $K$ show comparatively lower performance. The middle transformer blocks yield the most discriminative representations, highlighting the importance of both feature and block selection for optimal performance. Additional experimental results 
    on other datasets 
    are provided in the Appendix 
    \S S7.}
    \vspace{-4mm}
    \label{fig:block_feature}
\end{figure}

\noindent\textbf{Impact of Feature Selection.} 
We first analyze the impact of feature selection on the performance of HFR at a fixed timestep. Since the MMDiT block is a transformer-based layer (see \S\ref{sec:problem-form}), a natural choice for feature selection includes Query ($Q$), Key ($K$), and Value ($V$). Additionally, we consider the output of the attention layer ($A$) and the final output of the MMDiT block ($O$) as potential candidate features.
To investigate their effectiveness, we extract them 
at dataset-specific timesteps: $t=50$ for CUB, $t=100$ for Oxford Flowers and Stanford Cars, and $t=1$ for Aircraft. 
For consistency, we adopt the same timestep configuration in all subsequent ablation studies.
For the Oxford Flowers dataset (see Fig.~\ref{fig:block_feature}), \( Q \) achieves the highest accuracy at 90.6\%, followed by \( V \), which attains 88.7\%. The remaining features (\( K, A, \) and \( O \)) yield lower accuracies. A similar trend is observed across CUB, Aircraft, and Stanford Cars (see Appendix \S S7). 
These findings demonstrate that feature selection from different DiT components has a significant impact on the A-SelecT  discriminative  performance. 
Supplemented by the results presented in Appendix \S S6, 
both \( Q \) and \( V \) achieve comparable discriminative performance, but in most cases, \( Q \) outperforms \( V \). To ensure consistency and maximize performance, we adopt \( Q \) as the default feature throughout all experiments.

\noindent\textbf{Impact of Block Selection.} 
As introduced in \S\ref{sec:exp setup}, we adopt SD 3.5, a DiT-based diffusion model with 24 MMDiT blocks. Since the choice of block can strongly influence discriminative performance, we investigate the effect of selecting different blocks for feature extraction. We train individual classifiers using features from different blocks to assess their discriminative effectiveness. Results in Fig.~\ref{fig:block_feature} indicate that the most effective features come from a middle layer. This observation is consistent with findings of~\cite{kim2025autoregressive, hatamizadeh2024diffit}, which explains that early blocks primarily capture coarse information, while later blocks focus on fine details. The middle layers thus combine both types of information, making them more critical for discrimination. We further discuss 
fusing multiple blocks' features in Appendix \S S5.

\begin{table}[b]
\vspace{-4mm}
\begin{adjustbox}{width=0.75\width,center}

\begin{tabular}{l||cccc}
\hline \thickhline
\rowcolor{mygray}
Input Resolution & Oxford Flowers&CUB &Aircraft &Stanford Cars \\
\hline \hline

$256$ & 85.3\%  &  69.3\% &  71.5\%  &  85.7\%\\
$512$ & \textbf{90.6\%}  &  \textbf{78.6\%} &  \textbf{77.5\%}  &  \textbf{86.1\%}\\
$1,024$ & 81.7\%  &  66.2\%  & 73.6 \%  &  82.7\%\\
\hline
\end{tabular}
\end{adjustbox}

\caption{\textbf{Impact of Input Resolution} ranging from $256$ to $1,024$. An input size of $512$ yields the highest accuracies.
}
\label{table:table_inputsize_sel}

\end{table}

\noindent\textbf{Impact of Input Resolution.}
DiT possesses the advantageous property of accommodating variable input resolutions. We thus explore whether the input resolution of an image would influence DiT's discriminative performance. Specifically, we extract $Q$ from the 9-$th$ block with different input sizes (\ie, $256$, $512$, and $1,024$). As seen in Table~\ref{table:table_inputsize_sel}, we stop at the input size of $512$ because performance saturation is observed around this point. Further enlarging the resolution would result in a decrease in performance (\eg, 90.6\% $vs.$ 81.7\% on Oxford Flowers). We argue that this may be due to overparameterization~\cite{ jia2022visual, wang2024m, han20232vpt}.

\begin{figure}[t!]
    \centering
    \includegraphics[width=0.47\textwidth]{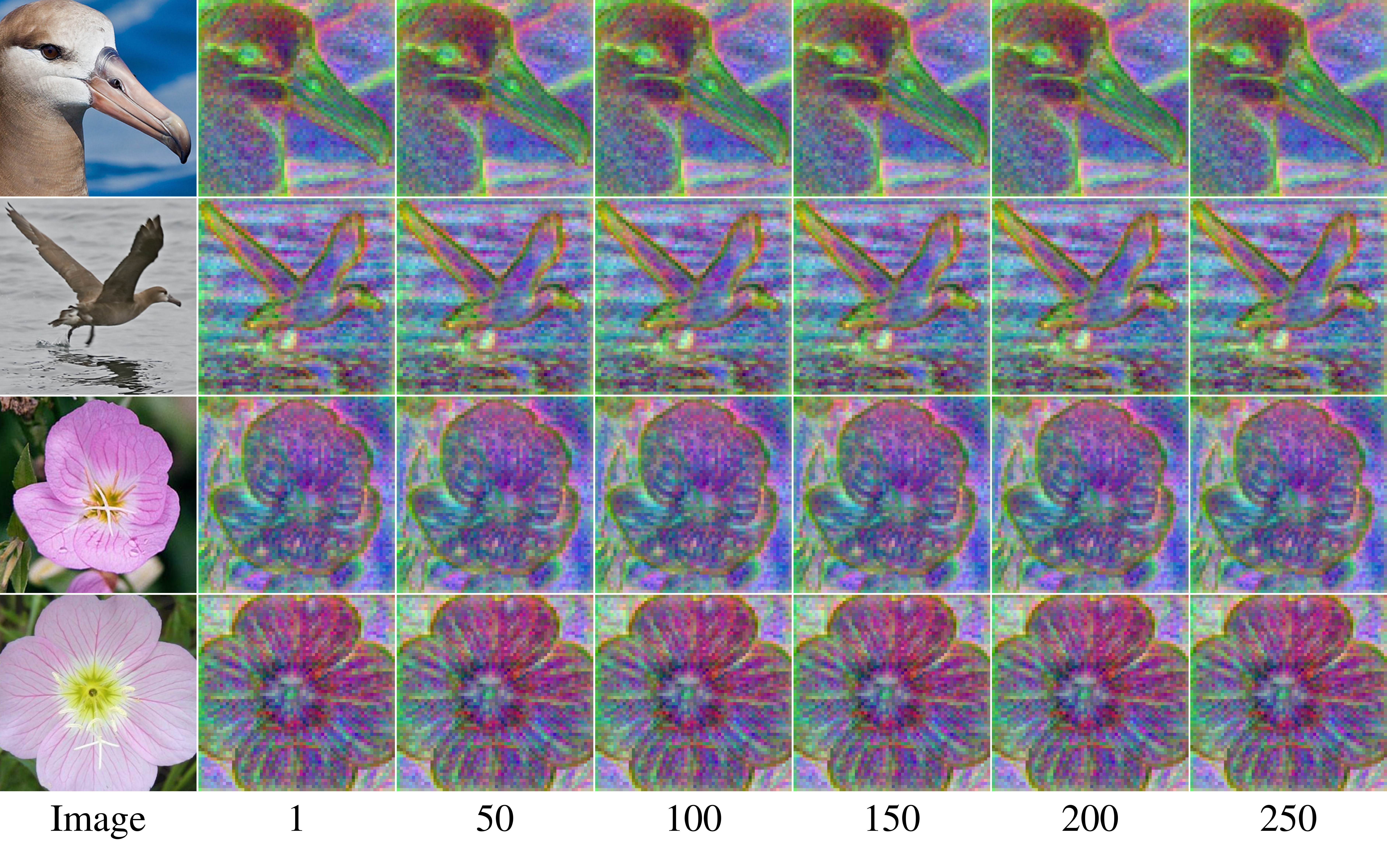}
    \vspace{-2mm}
    \caption{\textbf{Feature Visualization from CUB and Oxford Flowers.} Visualizations of four feature sets from the 9-$th$ block of SD 3.5. Each column represents $Q$ feature extracted at timestep $t \in \{1, 50, 100, 150, 200, 250\}$. Notably, the differences among features from various timesteps are subtle and challenging to discern visually, showing that manually selecting the discriminative feature extraction is impractical.
    More examples in Appendix \S S8.}
    \vspace{-4mm}
    \label{fig:feature_vis}
\end{figure}

\section{Discussions on A-SelecT}
\label{sec:discussion}

While the above results highlight the robustness of A-SelecT across discriminative tasks, it naturally raises the question of the \textit{necessity} of incorporating A-SelecT into diffusion model feature selection.
To investigate this, we conduct feature visualization and traversal search, adhering to recent advancements in diffusion models~\cite{mukhopadhyay2024text, meng2025not, chadebec2025lbm}, applied to SD 3.5 to ensure a fair comparison.
All experiments are conducted on CUB and Oxford Flowers datasets.
\subsection{DiT with Feature Visualization}\label{subsec:dit_with_feature_vis}
We follow \cite{meng2025not} and extract features at multiple timesteps: {1000, 950, 900, 850, 800, 700, 650, 600, 550, 500, 450, 400, 350, 300, 250, 200, 150, 100, 50, 1}. We then visualize and subjectively select (\ie, manual selection, following \cite{meng2025not}) timestep 1 for CUB and 250 for Oxford Flowers, respectively, as they appear to contain the most discriminative information. Feature visualization results are presented in Fig.~\ref{fig:feature_vis}. The results, as depicted in Table~\ref{table:table_visual}, indicate that the timesteps selected through the visualization strategy do not yield reliable outcomes. For example, we notice a substantial performance gap when compared to our approach on CUB (\ie, 72.3\% $vs.$ 78.6\%).
Moreover, manually comparing image visualizations across different timesteps proves to be both labor-intensive and challenging, significantly impeding efficient analysis. 
Consequently, we argue that feature visualization is both impractical and inefficient for diffusion model discriminative feature selection.

\begin{table}[t]
    \centering
    
    \begin{subtable}{\linewidth}
        \centering
        
\begin{adjustbox}{width=0.8\width,center}
\begin{tabular}{l||cc}
\hline \thickhline
\rowcolor{mygray}
Method & Oxford Flowers&CUB \\
\hline \hline

DiT-Visualization & 83.0\%  &  72.3\% \\
Ours & \textbf{90.6\%}  &  \textbf{78.6\%} \\

\hline
\end{tabular}
\end{adjustbox}
 \vspace{0.1cm}
\caption{\textbf{Comparision between A-SelecT and Feature Visualization.} }
\label{table:table_visual}
    \end{subtable}

    \vspace{0.2cm} 

    \begin{subtable}{\linewidth}
        \centering
        
\begin{adjustbox}{width=0.65\width,center}
\begin{tabular}{l||cc|cc|c}
\hline \thickhline
\rowcolor{mygray}
 & \multicolumn{2}{c|} {Oxford Flowers}& \multicolumn{2}{c|}{CUB}& GPU \\
\rowcolor{mygray}
\multirow{-2}{*}{Phase}& Ours & DiT-Traversal Search &  Ours & DiT-Traversal Search & hours \\

\hline \hline

Tuning        & 0.8 hrs  &  16.8 hrs  & 2.2 hrs &  47.0 hrs &  ${\sim}21\times$\\
A-SelecT & 0.6 hrs  &  0 hrs     & 1.6 hrs &  0 hrs&  -\\
\hline
Total      & 1.4 hrs  & 16.8 hrs  & 3.9 hrs &  47.0 hrs &  ${\sim}12\times$\\
 
\hline
\end{tabular}
\end{adjustbox}
 \vspace{0.1cm}
\caption{\textbf{Comparison between A-SelecT and Traversal Search.} The table presents the total time (in GPU hours) required by A-SelecT and traversal search to identify the optimal timestep $\hat{t}$. 
        }
\label{table:table_travers_time}
    \end{subtable}

\caption{\textbf{Discussions on A-SelecT} $ w.r.t$ feature visualization and traversal search further confirm the significance of our approach, showing outstanding automatic discriminative feature selection and training efficiency.}
\vspace{-4mm}
\end{table}

\subsection{DiT with Traversal Search}\label{subsec:dit_with_traversal_search}
In traversal search, we exhaustively train separate models at different timesteps: {1000, 950, 900, 850, 800, 700, 650, 600, 550, 500, 450, 400, 350, 300, 250, 200, 150, 100, 50, 1}. 
As introduced in \S\ref{sec:opt_sel}, A-SelecT involves computing HFR values for different timesteps, and selecting the timestep with the highest value for feature extraction. We only need to train the discriminative head in a single trial. 
In contrast, traversal search requires iterative training of the discriminative head until the optimal timestep is identified.
This increased granularity is directly associated with an
extended duration of training time, in this case, ${\sim}21\times$ when compared to our A-SelecT (see Table~\ref{table:table_travers_time}). The results demonstrate that our method is ${\sim}12\times$ faster than traversal search, which significantly enhances the efficiency of the 
feature selection. Notably, the efficiency advantage becomes even more pronounced with extended training. 
\section{Conclusion}
\label{sec:conclusion}
DiT has recently emerged as a promising alternative to traditional U-Net-based diffusion models for representation learning. However, its potential for discriminative tasks remains largely underexploited due to two core limitations: the lack of principled timestep selection and insufficient analysis of DiT’s internal representations. To address these challenges, we propose A-SelecT, a novel automatic timestep selection framework for effective and efficient DiT representation learning. Comprehensive experiments demonstrate that A-SelecT is able to: \textbf{I.} 
significantly optimize training schedules for DiT representation learning; and 
\textbf{II.} ensure peak performance among competitive methods. 
We posit that our research constitutes a foundational contribution to diffusion model representation learning.

\subsubsection*{Acknowledgments}
This research was supported by the National Science Foundation under Grant No. 2450068. This work used NCSA Delta GPU through allocation CIS250460 from the Advanced Cyberinfrastructure Coordination Ecosystem: Services \& Support (ACCESS) program, which is supported by U.S. National Science Foundation grants No. 2138259, No. 2138286, No. 2138307, No. 2137603, and No. 2138296.

The views and conclusions contained herein are those of the authors and should not be interpreted as necessarily representing the official policies or endorsements, either expressed or implied, of the U.S. Naval Research Laboratory (NRL) or the U.S. Government.

{
    \small
    \bibliographystyle{ieeenat_fullname}
    \bibliography{main}
}
\clearpage
\setcounter{page}{1}
\maketitlesupplementary

\maketitle
The supplementary is organized as follows: 

\renewcommand{\thesection}{S\arabic{section}}
\renewcommand{\thetable}{S\arabic{table}}
\renewcommand{\thefigure}{S\arabic{figure}}
\setcounter{table}{0}
\setcounter{figure}{0}
\setcounter{section}{0}

\begin{itemize}
  \item \S\ref{Appendix:datasets} details the \textbf{Datasets} utilized in the study, including statistical descriptions of each dataset.
  \item \S\ref{Appendix:pre_reults} offers \textbf{Additional Preliminary Results}.
  \item \S\ref{Appendix:impl_details} provides more \textbf{Implementation Details}.
  \item \S\ref{Appendix:observation-support} offers \textbf{Additional Visualizations} of feature decomposition.
  \item \S\ref{Appendix:ablation} presents new \textbf{Discussions on Feature Fusion}.
  \item \S\ref{Appendix:ablation_qv} presents \textbf{Discussions on Feature Selection between Query and Value}.
  \item \S\ref{Appendix:ablation_block_qkv} provides \textbf{Additional Results on Impact of Feature and Block Selection}.
  \item \S\ref{Appendix:feat_vis} offers \textbf{Additional Feature Visualization Examples}.
  \item \S\ref{Appendix:feat_vis_dit} presents \textbf{Visualization Feature Selection Results}.
  \item \S\ref{Appendix:res_to_HFR} presents \textbf{Discussions on Impact of Resolution on HFR}.
  \item \S\ref{Appendix:ddae_fgvc} provides \textbf{DDAE Classification Results on FGVC}.
  \item \S\ref{Appendix:HFR} provides \textbf{Additional Details on HFR}.
  \item \S\ref{Appendix:fisherscore} provides \textbf{Additional Details on Fisher Score}.
  \item \S\ref{Appendix:hfr_dits} provides \textbf{Additional HFR Results across multiple DiT Models}.
  \item \S\ref{Appendix:Discussion} gathers \textbf{Additional Discussions} on license, reproducibility, technical contributions, social impact and limitations, and future work.
\end{itemize}

\section{Datasets}
\label{Appendix:datasets}
In Table~\ref{table:app_datasets}, we provide statistical information of FGVC benchmark (\ie, Caltech-UCSD Birds (CUB)~\cite{wah2011caltech}, Aircraft~\cite{maji2013fine}, Stanford Cars~\cite{krause20133d}, NABirds~\cite{van2015building}, Stanford Dogs~\cite{khosla2011novel}, Oxford Flowers~\cite{nilsback2008automated}), ImageNet~\cite{deng2009imagenet}, and ADE20K~\cite{zhou2017scene}.

\begin{table}[ht]

\vspace{-3mm}

\begin{adjustbox}{width=0.80\width,center}
\begin{tabular}{l||ccc}
\hline \thickhline
\rowcolor{mygray}
Dataset & Class Number &Training Number &Test Number\\
\hline \hline

Aircraft &100  &6,667  & 3,333\\
Stanford Cars &196  &8,144  &8,041\\
CUB &200  &5,994  &5,794\\
Stanford Dogs &120  &12,000  &8,580\\
Oxford Flowers &102 &2,040 &6,149\\
NABirds &555  &23,929  &24,633\\
\hline
ImageNet   &1,000  &1.28M  &50,000\\
\hline
ADE20K    &150  &2,0210  &2000\\
    
\hline
\end{tabular}
\end{adjustbox}
\caption{\textbf{Datasets Statistical Details.}}
\label{table:app_datasets}
\end{table}

\section{Additional Preliminary Results}
\label{Appendix:pre_reults}
\begin{figure}[t!]
    \centering
    \includegraphics[width=0.5\textwidth]{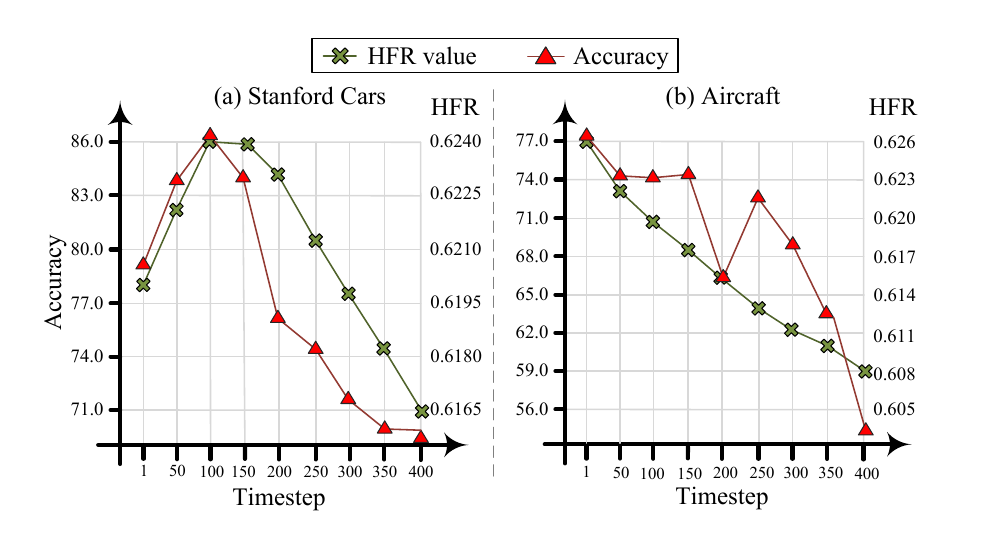}
    \vspace{-8mm}
    \caption{\textbf{More Preliminary Results} on the impact of the High-Frequency Ratio (HFR) with classification performance on Stanford Cars (a) and Aircraft (b).  
    }
    \vspace{-5mm}
    \label{fig:HFR2}
\end{figure}

We present additional preliminary results illustrating the relationship between HFR and classification accuracy on the Stanford Cars and Aircraft datasets, as shown in Fig.~\ref{fig:HFR2}. The results indicate a clear positive correlation: classification accuracy is consistently highest at the timestep where HFR reaches its maximum. Similar trends are observed on CUB and Oxford Flowers, further supporting HFR as a reliable and robust indicator of discriminative feature quality.

\section{Additional Implementation Details}
\label{Appendix:impl_details}
Here we provide more implementation details for our experiments. SD 3.5~\cite{esser2024scaling} released Large version model with 8 billion weights, while Medium version with 2 billion weights. In our experiment, we use the SD 3.5 Medium, comprising $24$ MMDiT blocks and features a backward process with 1,000 time steps for denoising. 
Given that SD 3.5 operates as a text-to-image model, we standardized the text condition to an empty string, ensuring uniformity across all features extraction. For text encoding, we exclusively employed the CLIP-G/14 encoder~\cite{radford2021learning}, electing not to utilize the CLIP-L/14~\cite{radford2021learning} or T5 XXL text encoders~\cite{raffel2020exploring}.
For a fair comparison, we include SDXL~\cite{meng2025not}) as a baseline, utilizing its U-Net backbone with 2.6 billion weights. This model is trained with textual conditioning, consistent with our SD 3.5 backbone. For the classification task, we report SDXL results under the same setting as our method. We also use an empty string as the text condition for SDXL.
For the ImageNet classification task, we do not report results for ResNet-50, as publicly available pretrained ResNet-50 checkpoints are mostly trained on ImageNet. Utilizing these models for evaluation will lead to an unfair comparison.

\section{Additional Results on High-Frequency Components}
\label{Appendix:observation-support}
\begin{figure}[t]
    \centering
    \includegraphics[width=0.48\textwidth]{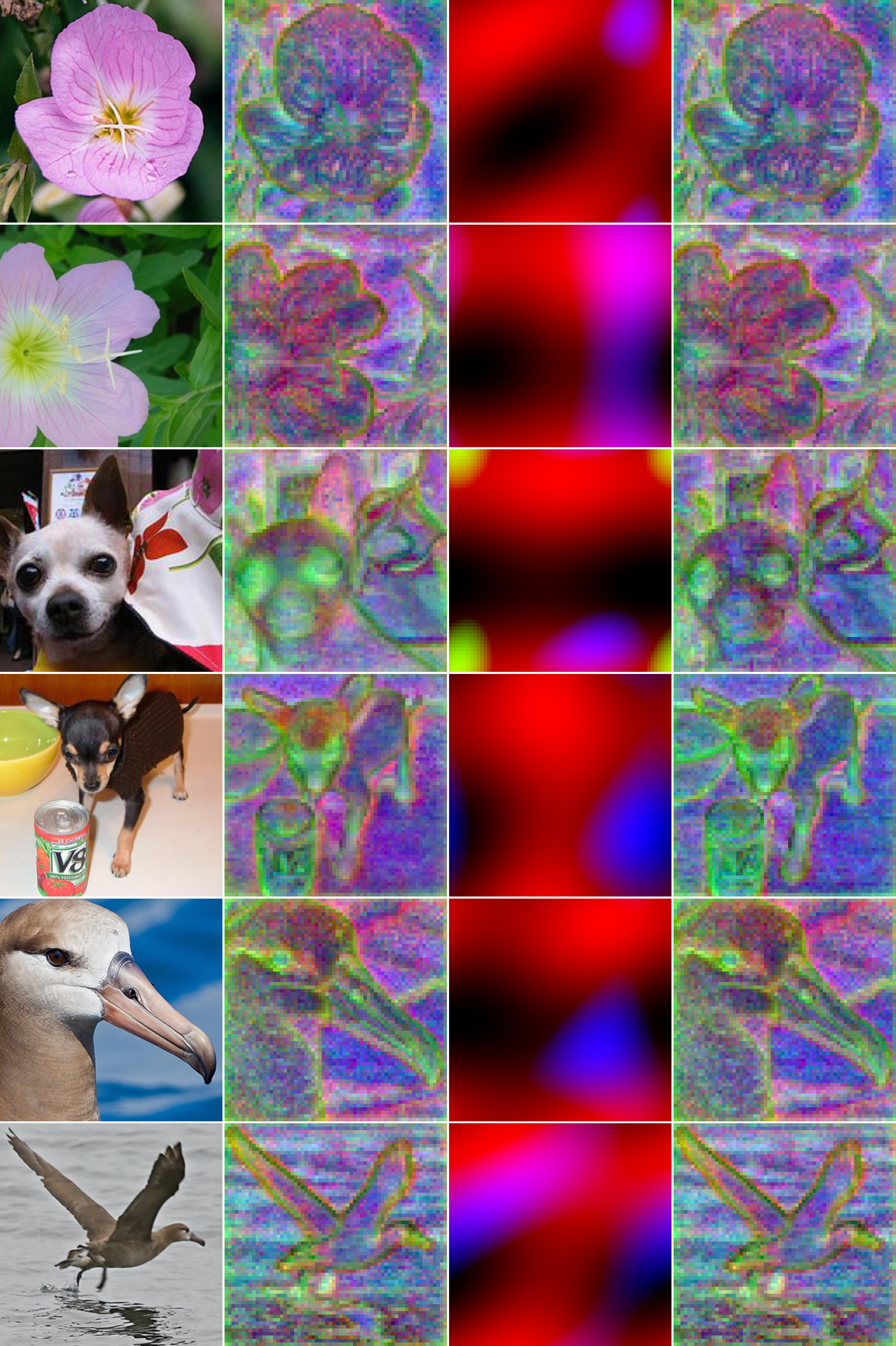}
    \put(-218,-9){\small Image}
    \put(-165,-9) { \small Original}
    \put(-105,-9){\small Low-freq}
    \put(-48,-9){\small High-freq}
    \vspace{-2.5mm}
    \caption{\textbf{More Visualization of Feature Decomposition Examples from CUB, Oxford Flowers and Stanford Dogs}. We present six sets of decomposition of the original features extracted from SD 3.5 into components that exclusively contain high-frequency and low-frequency information.
    }
    \vspace{-3mm}
    \label{fig:high_low_vis2}
\end{figure}

In Fig.~\ref{fig:high_low_vis2}, we provide more visualization results by decomposing of the original extracted features into components that exclusively contain high-frequency and low-frequency information. As seen, high-frequency features turn to contain more discriminative information, which is consistent with our observation in the main paper \S 3.3.

\begin{figure*}[t]
    \centering
    \includegraphics[width=0.95\textwidth]{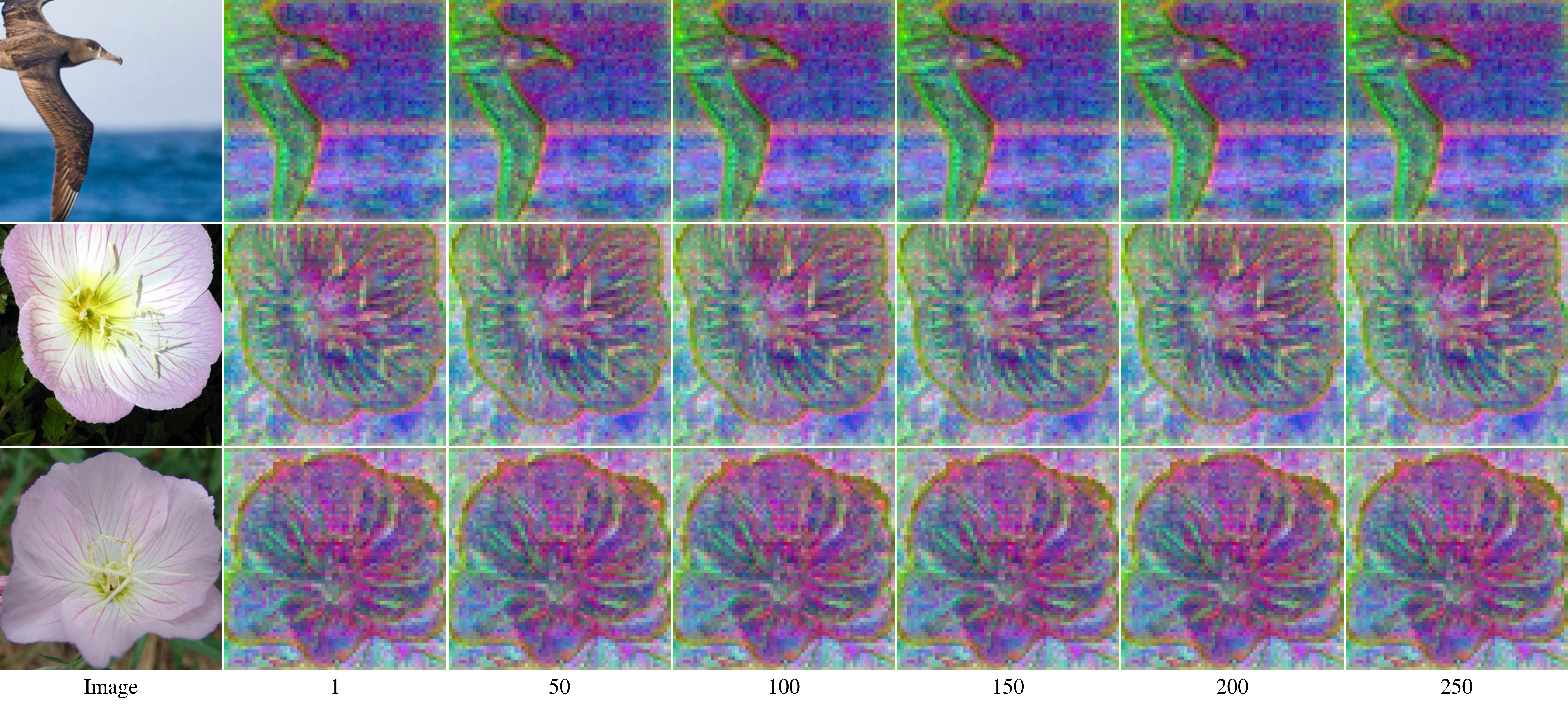}
    \vspace{-2.5mm}

    \caption{\textbf{More Feature Visualization Examples from CUB and Oxford Flowers.} We present visualizations of three feature sets extracted from the 9-$th$ block of SD 3.5. The columns represent features extracted at different timesteps. 
    }
    \vspace{3mm}
    \label{fig:more_feature_vis}
\end{figure*}

\section{Discussions on Feature Fusion}
\label{Appendix:ablation}
One major research question that may arise is whether we can achieve better performance when 
utilizing features from multiple blocks or different timesteps. We thus conduct an experiment on CUB dataset in Table~\ref{table:table_app_abl}.
The results indicate that incorporating additional features does not enhance accuracy. Even worse, the additional operations deteriorate our model's performance. This decrease in performance can be attributed to the increased complexity introduced to the model. We propose that the augmented complexity burdens the downstream classifier, thereby impairing its effectiveness.

\begin{table}

\begin{adjustbox}{width=0.85\width,center}
\begin{tabular}{cc||cc}
\hline \thickhline
\rowcolor{mygray}
Block & Timestep & CUB~\cite{wah2011caltech} \\
\hline \hline
9-$th$ + 6-$th$   &  50 & 71.9\% \\
9-$th$ + 12-$th$ &  50  &  72.5\% \\
9-$th$ &  50 + 1 &   76.1\%\\  
9-$th$ &  50 + 100  &  78.0\% \\
9-$th$ &  50  &  \textbf{78.6\%} \\

\hline
\end{tabular}
\end{adjustbox}
\caption{\textbf{Impact of Feature Fusion}. We extract features from different blocks at the same timestep and from the same block at various timesteps. The results indicate that the addition of more features does not lead to an improvement in performance; rather, it may actually decrease performance.
}
\label{table:table_app_abl}
\vspace{-1mm}
\end{table}

\begin{table}

\begin{adjustbox}{width=0.85\width,center}
\begin{tabular}{c||ccc|ccc}
\hline \thickhline
\rowcolor{mygray}
 & \multicolumn{3}{c|}{Aircraft} & \multicolumn{3}{c}{CUB} \\
\rowcolor{mygray}
Timestep & Query & Value& Key & Query & Value& Key \\
\hline \hline

1    & 77.5\% & 76.6\% & 72.1\% & 72.3\% & 71.3\%  & 47.5\%\\
50   & 74.3\% & 73.4\% & 69.7\% & 78.6\% & 76.1\%  & 68.2\%\\
100  & 74.2\% & 74.0\% & 70.7\% & 77.9\% & 73.1\%  & 64.2\%\\
150  & 74.6\% & 74.9\% & 62.0\% & 75.0\% & 71.0\% & 56.7\% \\
200  & 66.5\% & 72.0\% & 48.7\% & 72.2\% & 68.7\% & 62.0\% \\
250  & 72.5\% & 55.6\% & 59.6\% & 67.3\% & 66.0\% & 63.3\% \\
300  & 68.9\% & 47.2\% & 59.8\% & 63.9\% & 63.8\% & 56.7\% \\
350  & 63.8\% & 62.6\% & 56.8\% & 64.8\% & 43.8\% & 58.3\% \\
400  & 53.5\% & 54.0\% & 37.8\% & 63.6\% & 52.0\% & 41.8\% \\
450  & 43.8\% & 55.2\% & 56.1\% & 57.7\% & 17.3\% & 35.5\% \\
500  & 45.9\% & 49.5\% & 28.9\% & 37.2\% & 40.5\% & 35.8\% \\
550  & 30.1\% & 21.3\% & 30.0\% & 27.2\% & 33.5\% & 26.3\% \\
600  & 35.7\% & 17.5\% & 23.5\% & 27.5\% & 25.2\% & 24.5\% \\
650  & 25.8\% & 20.5\% & 10.3\% & 19.8\% & 16.5\% & 17.2\% \\
700  & 15.3\% & 9.7\% & 6.3\% & 15.5\% & 13.0\% & 7.2\% \\
750  & 12.0\% & 9.9\% & 8.1\% & 12.3\% & 7.8\% & 5.3\% \\
800  & 6.6\% & 5.1\% & 6.7\% & 8.0\% & 3.7\% &2.1\% \\
850  & 5.8\% & 3.0\% & 4.8\% & 4.2\% & 3.5\% & 3.5\%\\
900  & 3.4\% & 3.3\% & 3.7\% & 2.0\% & 3.5\% & 1.9\% \\
950  & 2.2\% & 2.5\% & 2.7\% & 1.3\% & 1.7\% & 1.2\% \\
1000  & 1.3\% & 1.5\% & 1.6\% & 0.8\% & 0.8\% & 0.6\% \\
\hline
\end{tabular}
\end{adjustbox}
\caption{\textbf{Accuracies of Query, Value, and Key Features Across Timesteps}, respectively. We report top-1 classification accuracy using query, value and key features at various diffusion timesteps on the Aircraft and CUB datasets. While query and value features show similar performance overall, query features outperform in a greater number of cases.}

\label{table:ablation_qv}
\vspace{-3mm}
\end{table}

\section{Discussions on Feature Selection}
\label{Appendix:ablation_qv}
Table \ref{table:ablation_qv} presents the classification accuracies of query, value, and key features on the CUB and Aircraft across different timesteps, respectively. We observe that both query and value features achieve comparable performance, substantially outperforming the key feature. Notably, the query feature generally outperforms the value feature in most cases. Based on this observation, we adopt the query feature for DiT representation learning for all experiments.

\section{Additional Results on Impact of Feature and Block Selection}
\label{Appendix:ablation_block_qkv}
\begin{figure*}[t!]
    \centering
    \vspace{-10mm}
    \includegraphics[width=1.00\textwidth]
    {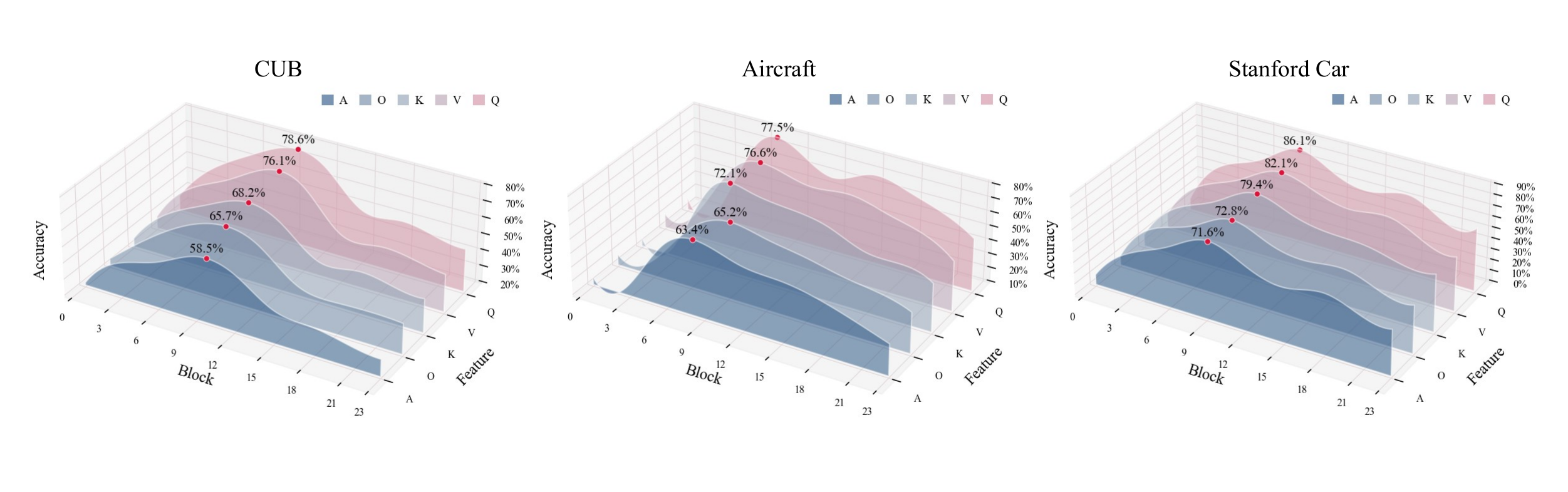}
    \vspace{-17mm}
    \caption{\textbf{Impact of Feature and Block Selection.} We show more block and feature performance on CUB, Aircraft, and Stanford Car. The figure shows the consistent result with Oxford Flower dataset.}
    \label{fig:block_feature_cub_air_car}
\end{figure*}

In Fig.~\ref{fig:block_feature_cub_air_car}, we provide additional results analyzing the impact of block and feature selection across multiple datasets, including CUB, Aircraft, and Stanford Cars. Consistent to the results shown in our main paper, the block and feature influence the performance. Consistent with the main paper (\S 4.3), both factors strongly influence representation's discriminative performance. Among transformer block components, $Q$ features achieve the highest accuracy, followed by $V$, while $K$, $A$, and $O$ perform worse. For block selection, features from middle layers consistently outperform those from early or late layers, as they capture a balanced mix of coarse and fine-grained representations. These findings confirm that optimal feature and block choices are crucial for maximizing discrimination.

\section{Additional Feature Visualization Examples}
\label{Appendix:feat_vis}
In Fig.~\ref{fig:more_feature_vis}, we provide more visualization examples of features extracted at different timesteps. Consistent to the results shown in our main paper, manual
selections of discriminative features at different timesteps are impractical and ambiguous.

\begin{figure*}[t]
    \centering
    \includegraphics[width=1\textwidth]{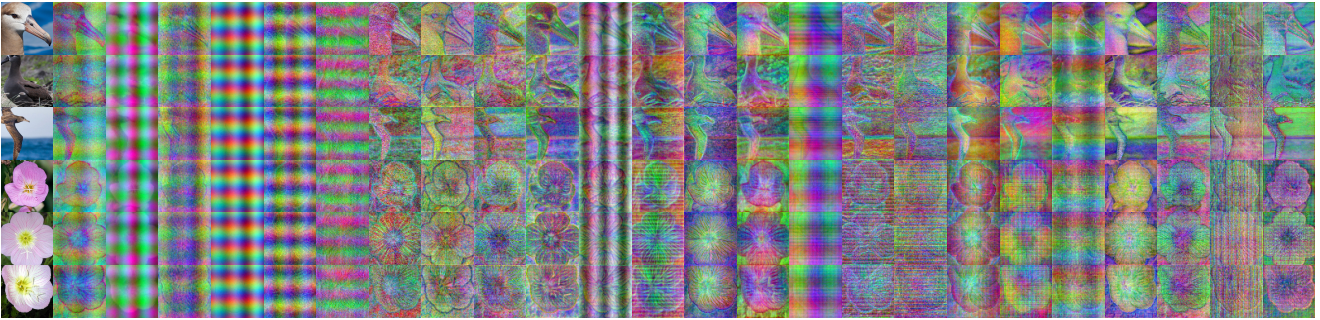}
    \vspace{-2.5mm}

    \caption{\textbf{Feature Visualization Across Blocks on CUB and Oxford Flowers.} The first column shows original input images and subsequent columns presents feature visualizations extracted from the 0-$th$ to 23-$th$ block of SD 3.5. The top three rows correspond to samples from CUB and the bottom three rows are from Oxford Flowers.
    }
    \vspace{3mm}
    \label{fig:blocks_feature_vis}
\end{figure*}

\section{Visualization Feature Selection Results}
\label{Appendix:feat_vis_dit}
We exam the visualization feature selection method from~\cite{meng2025not} further for DiT block selection on the CUB and Oxford Flowers datasets, visualizing features at timestep 50 for CUB and timestep 100 for Oxford Flowers in Fig.~\ref{fig:blocks_feature_vis}. Based on visualization, we manually identify the most informative features from blocks 7, 0, 20, and 12. These selected features are subsequently combined and evaluated on downstream tasks. However, Table~\ref{table:table_vis_dit} shows that these selected features underperform compared to the single feature from block 9. This suggests that the visualization selection method is not effective for the DiT model.

\begin{table}[t]
\vspace{-3mm}

\begin{adjustbox}{width=0.85\width,center}
\begin{tabular}{c||cc}
\hline \thickhline
\rowcolor{mygray}
Block  & CUB & Oxford Flowers \\
\hline \hline
7-$th$    & 66.2\%        & 52.0\%\\
7-$th$ + 0-$th$   & 69.7\%        & 84.3\%\\
7-$th$ + 0-$th$ + 20-$th$  & 66.2\%      & 67.2\%\\  
7-$th$ + 0-$th$ + 20-$th$ + 12-$th$ & 58.5\%       & 73.0\% \\

\hline
\end{tabular}
\end{adjustbox}
\caption{\textbf{Visualization Feature Selection Results on
DiT}. 
}
\label{table:table_vis_dit}
\vspace{-3mm}
\end{table}

\begin{table}[t]
\vspace{-3mm}
\begin{adjustbox}{width=0.83\width,center}
\begin{tabular}{c||cc|cc|cc}
\hline \thickhline
\rowcolor{mygray}
 & \multicolumn{2}{c|}{256} & \multicolumn{2}{c|}{512} & \multicolumn{2}{c}{1024} \\
\rowcolor{mygray}
Timestep & \texttt{Acc.} & HFR  & \texttt{Acc.} & HFR & \texttt{Acc.} & HFR\\
\hline \hline

1    & \textbf{71.5\%} &\textbf{0.6163} & 72.3\% &0.6199  & 65.2\% & 0.6180\\
50   & 69.3\% &0.6131 & \textbf{78.6\%} &\textbf{0.6223}  & 66.2\% & 0.6195\\
100  & 68.2\% &0.6110 & 77.9\% &0.6221  & 67.8\% & 0.6208\\
150  & 62.0\% &0.6092 & 75.0\% &0.6217  & \textbf{74.8\%} & \textbf{0.6234}\\
200  & 61.3\% &0.6077 & 72.2\% &0.6212  & 69.7\% & 0.6164 \\
250  & 57.8\% &0.6059 & 67.3\% &0.6200  & 72.4\% & 0.6144 \\
300  & 46.5\% &0.6045 & 63.9\% &0.6201  & 70.0\% & 0.6128 \\
350  & 34.0\% &0.6031 & 64.8\% &0.6186  & 63.3\% & 0.6111 \\
400  & 27.5\% &0.6020 & 63.6\% &0.6171  & 68.2\% & 0.6096 \\
450  & 26.2\% &0.6011 & 57.7\% &0.6155  & 53.3\% & 0.6082 \\
500  & 14.3\% &0.5998 & 37.2\% &0.6138  & 52.0\% & 0.6067 \\
550  & 14.5\% &0.5987 & 27.2\% &0.6123  & 51.7\% & 0.6055\\
600  & 11.7\% &0.5977 & 27.5\% &0.6112  & 46.2\% & 0.6048 \\
650  & 7.3\% &0.5962 & 19.8\% &0.6100   & 36.3\% & 0.6039 \\
700  & 5.7\% &0.5943 & 15.5\% &0.6090   & 38.7\% & 0.6034 \\
750  & 4.3\% &0.5916 & 12.3\% &0.6075   & 24.4\% & 0.6025 \\
800  & 3.5\% &0.5889 & 8.0\% &0.6063   & 16.8\% & 0.6019\\
850  & 2.0\% &0.5857 & 4.2\% &0.6042   & 11.1\% & 0.6000\\
900  & 1.7\% &0.5824 & 2.0\% &0.6013   & 8.2\% &  0.5959\\
950  & 1.2\% &0.5711 & 1.3\% &0.5946   & 2.7\% &  0.5851\\
1000  & 1.2\% &0.5720 & 0.8\% &0.5870  & 0.8\% & 0.5789\\
\hline
\end{tabular}
\end{adjustbox}
\caption{\textbf{Accuracies and HFR across Input Resolutions and Timesteps on CUB.} The highest classification accuracy consistently corresponds to the highest HFR value across input resolutions.}

\label{table:res_to_HFR}
\vspace{-3mm}
\end{table}

\section{Discussions on Impact of Resolution on HFR}
\label{Appendix:res_to_HFR}
In Table \ref{table:res_to_HFR}, we report the classification accuracies of query features on the CUB dataset across different input resolutions (\ie, $256$, $512$, and $1,024$) and timesteps (\ie, $1000$, $950$, $900$, $850$, $800$, $750$, $700$, $650$, $600$, $550$, $500$, $450$, $400$, $350$, $300$, $250$, $200$, $150$, $100$, $50$, $1$). The results indicate that the optimal timestep yielding the highest classification performance varies with the input resolution. Notably, these optimal timesteps consistently correspond to the highest HFR values, suggesting that HFR remains robust to changes in resolution. This implies that HFR effectively identifies the optimal timestep regardless of the input resolution.

\section{DDAE Classification Results on FGVC}
\label{Appendix:ddae_fgvc}
Denoising Diffusion Autoencoders (DDAE)~\cite{ddae2023} extracts layer-wise output features from Diffusion Transformer~\cite{peebles2023scalable}. We evaluate DDAE on FGVC datasets (\ie, Aircraft, Stanford Cars, CUB, Stanford Dogs, Oxford Flowers and NABirds) and compare it against our method. The results, in Table \ref{table:ddae_fgvc}, demonstrate that our method significantly outperforms DDAE across all datasets. For example, our method achieves \textbf{86.1\%} accuracy on Stanford Cars and \textbf{78.4\%} on NABirds, compared to 20.0\% and 17.1\% with DDAE, respectively. These consistent gains across datasets demonstrate the effectiveness of our approach.

\begin{table}[ht]

\begin{adjustbox}{width=1.0\width,center}
\begin{tabular}{l||ccc}
\hline \thickhline
\rowcolor{mygray}
Dataset & DDAE &Ours\\
\hline \hline
Aircraft &19.3\%  &\textbf{77.5\%}  \\
Stanford Cars &20.0\%  &\textbf{86.1\%}  \\
CUB &25.4\%  &\textbf{78.6\%} \\
Stanford Dogs &49.2\%  &\textbf{83.5\%} \\
Oxford Flowers &73.4\% &\textbf{90.6\%}  \\
NABirds &17.1\%  &\textbf{78.4\%}  \\
\hline
Mean &34.1\%  &\textbf{82.5\%}\\
\hline
\end{tabular}
\end{adjustbox}

\caption{\textbf{DDAE Classification Results on FGVC.}}
\label{table:ddae_fgvc}
\end{table}

\section{Additional Details on HFR}
\label{Appendix:HFR}
We compute HFR on the test dataset to ensure that it captures discriminative information from unseen data. The Gaussian high-pass filter threshold is set to 30.

\section{Additional Details on Fisher Score}
\label{Appendix:fisherscore}
\begin{figure}[b!]
    \centering
    \vspace{-2mm}
    \includegraphics[width=0.5\textwidth]
    {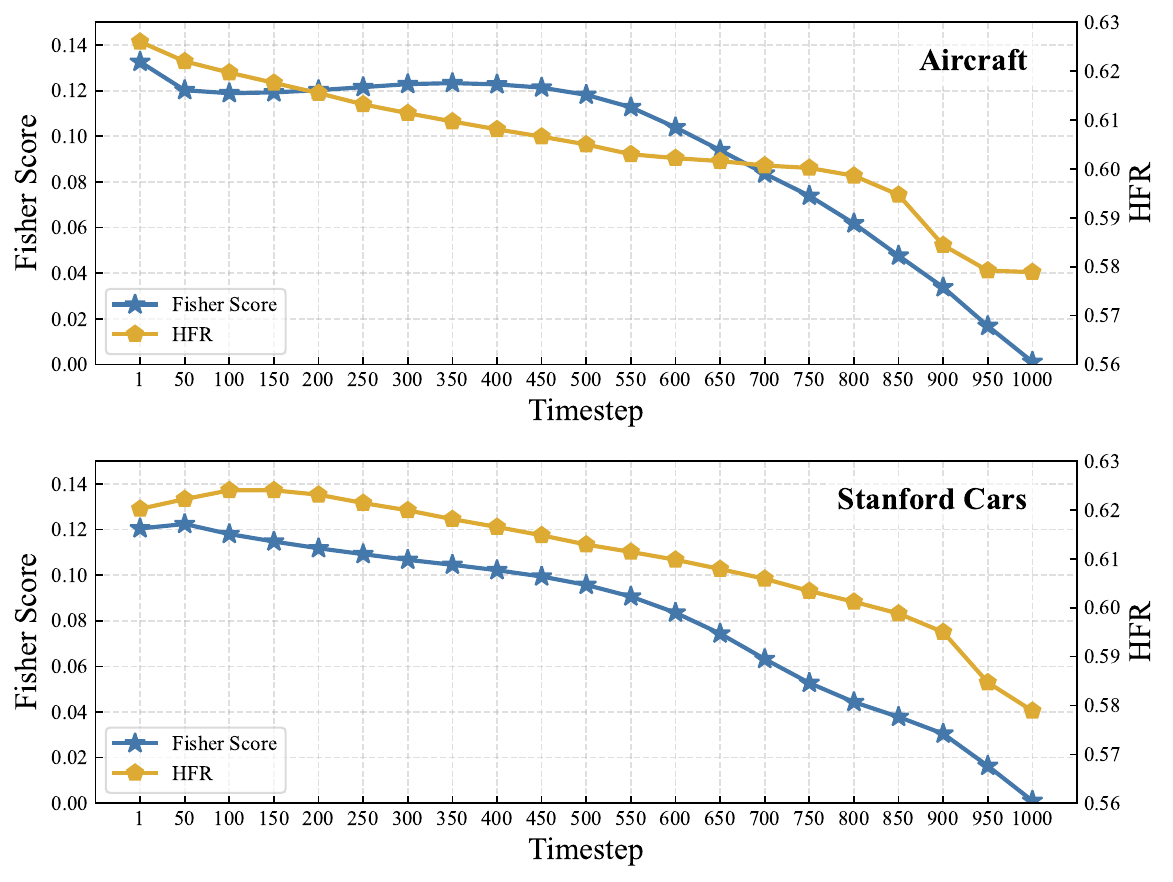}
    \caption{\textbf{More Comparison of HFR and Fisher Score} across timesteps on Aircraft (top) and Stanford Cars (bottom). The results show that HFR and Fisher Score exhibit consistent trends.}
    \vspace{-2mm}
    \label{fig:HFRvsFS_air_car}
\end{figure}

We compute the Fisher score on the test dataset same as HFR. To obtain a one-dimensional embedding for each sample, we apply mean pooling over the token dimension of the two-dimensional feature representations.
More results about relationship between Fisher Score and HFR are shown in Fig.~\ref{fig:HFRvsFS_air_car}

\section{Additional HFR Results across multiple DiT Models}
\label{Appendix:hfr_dits}
\begin{figure}[t!]
    \centering
    \includegraphics[width=0.5\textwidth]{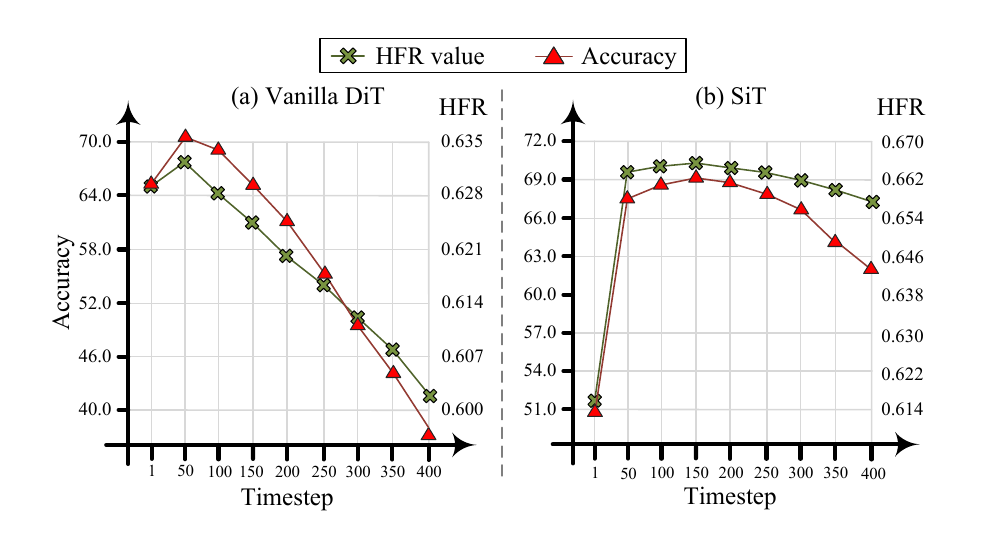}
    \vspace{-8mm}
    \caption{\textbf{Comparison of HFR and Classification Accuracy across multiple DiT Models} on Oxford Flowers. The alignment of peak accuracy with maximum HFR values demonstrates the consistent generalization of HFR.}
    \vspace{-5mm}
    \label{fig:dit_sit_hfr_acc}
\end{figure}

To further examine the generalization of our proposed HFR, we evaluate it on different DiT models, including Vanilla DiT~\cite{peebles2023scalable} and SiT~\cite{ma2024sit}, using Oxford Flowers dataset under the same experimental settings as in \S 4.1. As shown in Fig.~\ref{fig:dit_sit_hfr_acc}, HFR values exhibit strong alignment with classification accuracy across both models. The highest accuracy consistently appears at the timestep where HFR reaches its maximum, demonstrating the robustness and generalization of HFR across DiT models.

\section{Discussion}
\label{Appendix:Discussion}

\subsection{Asset License and Consent}
\label{Appendix:License}
Stable Diffusion 3.5 is licensed under https://
huggingface.co/stabilityai/stable-diffusion-3.5-large/blob\\
main/LICENSE.md

\subsection{Reproducibility}
\label{Appendix:reproduce} 
To guarantee reproducibility, our full implementation shall be publicly released upon paper acceptance. 
We provide the pseudo code of our proposed A-SelecT in Algorithm~\ref{alg:code}.

\begin{algorithm}[t!]
    \caption{Pseudo-code of A-SelecT in a PyTorch-like style.}
    \label{alg:code}
    \definecolor{codeblue}{rgb}{0.25,0.5,0.5}
    \lstset{
     backgroundcolor=\color{white},
     basicstyle=\fontsize{8pt}{8pt}\ttfamily\selectfont,
     columns=fullflexible,
     breaklines=true,
     captionpos=b,
     commentstyle=\fontsize{9pt}{9pt}\color{codeblue},
     keywordstyle=\fontsize{9pt}{9pt},
    }
   \begin{lstlisting}[language=python]
   # timesteps: timesteps used for computing HFR
   # epochs: number of training epochs
   # dif_model_path: diffusion model path

   def A-SelecT(timesteps, epochs, dif_model_path):
        pipe = StableDiffusion3Pipeline.from_pretrained(dif_model_path)
        downstream_network = setup_head()
        optimizer = AdamW(downstream_network.named_parameters())
        HFR_list = []
        
        #get optimal timestep
        for t in timesteps:
            HFR_t = compute_HFR(t, train_dataloader, pipe)
            HFR_list.append(HFR_t)
        optimal_t, highest_HFR = get_optimal_t(HFR_list)

        #train downstream network
        for epoch in range(epochs):
            for step, batch in enumerate(train_loader):
                features = pipe.transformer.extract(batch, optimal_t)
                output = downstream_network(features)
                loss = loss_fn(output, batch)
                loss.backward()
                optimizer.step()
                downstream_network.zero_grad()
       
   \end{lstlisting}
   \end{algorithm}

\subsection{Technical Contributions}
\label{Appendix:technical_contributions}

Our study presents three principal technical contributions: \textbf{First}, the inspiration for this research derives from the observation that high-frequency details, such as edges, textures, and corners, typically harbor more discriminative information. This insight has led to the development of the High-Frequency Ratio (HFR) metric. \textbf{Second}, a significant challenge in using diffusion models for extracting features is the selection of the most informative timestep from the extensive denoising trajectory. Traditional methods depend on exhaustive brute force searching or subjective manual selection, both of which are inefficient and potentially inaccurate. Our implementation of the HFR addresses this issue by providing a reliable and computationally efficient method for identifying the optimal timestep. \textbf{Third}, this paper is pioneering in its analysis of Diffusion Transformer (DiT)-based models for feature extraction. Through comprehensive experiments, we demonstrate that our approach not only overcomes the limitations of existing methods but also achieves state-of-the-art performance, substantiating the efficacy of DiT-based models as robust tools in representation learning.

\subsection{Limitations}
\label{Appendix:social_impact}
Although our HFR performs effectively for selecting a single feature within the DiT model, it remains unclear whether this approach is equally viable for simultaneously selecting multiple features. In \S\ref{Appendix:ablation}, we find that under the pipeline of A-SelecT, additional features from different blocks or timesteps do not lead to better performance. However, other quantitative metric might be suitable under such the scenarios. In this sense, further investigation is needed to ascertain the applicability and effectiveness of the HFR metric in involving multi-feature extraction from DiT models.

\subsection{Future Work}
\label{Appendix:future_work}
As discussed in \S\ref{Appendix:social_impact}, although our ablation study in \S\ref{Appendix:ablation} demonstrates that incorporating more features leads to diminished downstream performance, it is plausible that additional features could provide more discriminative information. The effective utilization of this increased information warrants further investigation. 
Furthermore, incorporating additional features introduces new challenges, including training efficiency and the precise identification of multiple discriminative feature candidates.

\end{document}